\def\arxivbuild{1}
\documentclass{article}

\usepackage{iclr2027_conference,times}
\usepackage{amsmath}
\usepackage{longtable}
\ifdefined\arxivbuild
  \iclrfinalcopy
\fi

\usepackage[utf8]{inputenc} 
\usepackage[T1]{fontenc}    
\usepackage{hyperref}       
\usepackage{url}            
\usepackage{booktabs}       
\usepackage{amsfonts}       
\usepackage{nicefrac}       
\usepackage{microtype}      
\usepackage{xcolor}         
\usepackage{graphicx}       
\usepackage{float}          
\usepackage{multirow}       
\title{Beyond Symmetric Agents:\\Cognitive Diversity and Multi-Agent Debate\\in Small Language Models}

\author{%
Leonardo Ferreira$^\dagger$, Gardenia Liu$^\ddagger$, Kaden Zheng$^\ddagger$
  \thanks{Corresponding author:$\,$leonardo.ferreira@childrens.harvard.edu.} \\
  $\dagger$ Boston Children's Hospital, Harvard Medical School\\
  $\ddagger$ Harvard School of Engineering and Applied Sciences, Harvard College
}

\begin{document}

\maketitle
\ifdefined\arxivbuild\lhead{Preprint}\fi

\begin{abstract}
  Multi-agent debate (MAD) reportedly improves reasoning and factuality
  over single-model inference, but prior work treats agents as symmetric
  peers, leaving open what drives the gains. We test the hypothesis that
  \emph{cognitive diversity} among agents is the driver, in the setting
  where the question is still measurable: small open-weight models with
  benchmark headroom. Across 23 models from eleven vendor families, five
  tasks, and 5,500+ debate and control runs, we vary diversity along three
  axes --- personas, sampling temperature, and model identity --- pairing
  every debate configuration with a generation-budget-matched majority-vote
  control. The hypothesis is rejected on every axis. Debate beats single-agent
  inference (3--7 points where tasks have headroom) but at matched budget conditions
  it ties or even loses to self-consistency sampling at 1.6$\times$ the wall-clock and 3.4$\times$   the token cost. Persona prompting reduces accuracy and a dose-response
  experiment over each model's full combinatorial persona space shows the
  cost is a \emph{persona} tax, not a \emph{diversity} tax: redundant
  personas hurt most, while maximally-diverse teams recover part of the loss.
  Furthermore, mixed-model teams \emph{lose} to majority votes over their own rosters, with
  accuracy tracking member capability rather than heterogeneity, and
  nearly all of debate's benefit comes from the first exchange of
  answers. We further identify a pervasive measurement hazard in which debate
  transcripts silently overflow serving context windows, whose
  correction alone moves our debate-versus-sampling comparison from
  $-1.8$ points to parity. Our results recast reported MAD gains as an
  ensemble-sampling effect and provide the budget-matched,
  contamination-checked baseline bar that future debate mechanisms
  should be required to clear.
\end{abstract}

\section{Introduction}

In multi-agent debate (MAD), multiple instances of language-model agents independently propose solutions to a given task, read one another's reasoning, and iteratively refine their answers; \citet{du2023improving} report significant gains over single-agent inference for GPT-3.5/4 on arithmetic, factual questions, and biography generation. Yet prior work treats \emph{agents} as identical copies of one model with a single prompt, observing performance gains that intuition says should require \emph{diversity}. This paper tests the hypothesis that cognitive diversity among agents drives debate's effectiveness, operationalized along three axes: persona diversity (distinct reasoning styles via system prompts), temperature diversity (distinct sampling temperatures), and model diversity (distinct base models).

Unlike prior work on large proprietary models, we experiment with small open-weight models: 23 models from eleven vendor families, 0.6B--32B dense plus a few sparse MoE variants (up to 109B total, 17B active). Open weight models were chosen because they allow for reproducibility and per-model persona optimization in each model's own \emph{representation space}; modest per-run compute enables a grid of more than 5{,}500 debate and control runs across five tasks (Appendix~\ref{app:design}); and, decisively, small models retain benchmark headroom, \emph{i.e.}, on saturated benchmarks a frontier model leaves debate nothing to improve, so ``does debate help?''\ is measurable only where models can still fail. Our pool spans solo accuracies from roughly 30\% to 92\%.

Our study returns a negative answer on every axis. MAD beats single-agent inference where tasks have headroom ($+3.3$ to $+6.6$ points), but matched self-consistency controls show that the gain is an \emph{ensemble-sampling} effect and that at equal generation counts, debate ties or loses to sampling-plus-voting on every task (Table~\ref{tab:main}) at $1.6\times$ the wall-clock and $3.4\times$ the total token count. Persona prompting reduces accuracy (Figure~\ref{fig:persona_by_task}), and a dose-response experiment over each model's full combinatorial persona space (an 11-rung determinant ladder from most redundant to most diverse team) shows that the ``tax'' is paid by wearing personas at all and not by their diversity. Redundant teams are the most harmful configuration we test, and accuracy is flat to mildly \emph{increasing} in diversity (Figure~\ref{fig:dose_response}). Heterogeneous model teams lose to majority votes over their own rosters ($-1.04$ points over 80 stratified teams, replicated at a second seed), with monotonic accuracy over member capability, not diversity (Table~\ref{tab:mixed}). Essentially all of multi-agent debate's benefit arrives with the first exchange of answers. Finally, we document a measurement hazard that inflated debate's apparent deficit before correction: multi-round transcripts silently overflow serving context windows for long-form reasoning models (Section~\ref{sec:ctx}).

\section{Related Work}

\paragraph{Multi-agent debate and its critiques.}
\citet{du2023improving} introduced multi-agent debate, reporting gains for GPT-3.5 agents on arithmetic, reasoning, and biography generation. Extensions include evaluation~\citep{chan2023chateval} and divergent-thinking prompting~\citep{liang2023encouraging}. A critical line has since questioned the mechanism of how and why this improvement has been observed: \citet{wu2025multi} argue the gains may be sampling effects and that at matched inference budgets, homogeneous debate frequently fails to beat self-consistency or majority voting~\citep{smit2024mad,zhang2025stop}, a pattern explained by modeling debate as a martingale over the initial answer distribution~\citep{choi2025debate}, while plain sampling-and-voting scales on its own~\citep{li2024more}. The most consistently positive reported reason is \emph{model heterogeneity}~\citep{chen2023reconcile,zhang2025stop}, and the scalable-oversight lineage finds debate's clearest wins under information asymmetry~\citep{khan2024debating,kenton2024scalable}. The sampling side of the comparison has meanwhile grown its own efficiency literature: adaptive stopping rules that cut samples without losing accuracy~\citep{aggarwal2023adaptive,li2024esc,wang2025dsc}, reliability-weighted accumulation~\citep{kim2026reasc}, and ranked-ballot aggregation~\citep{wang2025rankedsc}. These methods only strengthen the control debate must beat; and newer debate frameworks inject structural diversity by construction, via path allocation, process-level critique, and tool-based verification~\citep{li2026dynadebate}.

We aim to address the following gaps this literature leaves open: budget-matched debate-versus-sampling on small open-weight models across tasks; per-model, embedding-optimized persona selection with a full \emph{dose-response} over the combinatorial persona space; tier-stratified heterogeneous teams evaluated against voting ensembles built from the same rosters~\citep{wang2023self,dietterich2000ensemble}; and context-window ceilings measured as a confound rather than assumed away. Small open-weight models~\citep{team2024gemma,deepseek2024r1} make this goal measurable: they retain benchmark headroom and support thousands of matched controls. A reference-verified survey of the field (2023--2026) accompanies the paper as supplementary material.

\section{Methods}

\subsection{Multi-Agent Debate Protocol}

We adopt the protocol of \citet{du2023improving}: in round one, each of the $n$ agents answers independently, and in later rounds each agent sees all other agents' previous-round responses and may maintain or revise its answer. After $k$ rounds, final answers are aggregated by majority vote (discrete tasks) or judged independently (generation tasks).

\subsection{Diversity Conditions}

Our experiments manipulate cognitive diversity through four conditions, each evaluated against matched-budget controls. The homogeneous baseline replicates the standard multi-agent debate setup from prior work, where all agents use the same model; unless a condition specifies otherwise, agents receive mildly diversified sampling temperatures evenly spaced in $[0.5, 1.0)$, so that no condition conflates its treatment with trivially identical decoding.

For persona diversity, we wrote 100 reasoning personas in two 50-persona tiers: moderate professional styles (``meticulous analyst'') and extreme unconventional ones (``Zen master communicating in koans''). The personas were injected as system prompts. All persona teams in the paper are selected from the extreme tier, whose 50 personas appear verbatim in Appendix~\ref{app:personas}; the moderate tier served early calibration only. Persona \emph{teams} are not sampled arbitrarily: for each model and team size $n\in\{3,5,7\}$ we select the combination maximizing the diversity objective of Section~\ref{sec:divmetric} in that model's own representation space, and the dose-response experiment of Section~\ref{sec:persona} extends this to an 11-rung ladder spanning each model's full combinatorial range.

To disentangle deliberation from ensembling, every debate configuration is paired with a generation-budget-matched \emph{self-consistency} control~\citep{wang2022selfconsistency}: $N \in \{9, 15, 21\}$ independent single-round samples with majority voting, matching the total generations of $\{3,5,7\}$ agents $\times$ 3 rounds. Self-consistency runs come in two flavors --- uniform temperature (the control for persona debates) and the same temperature spread as debate teams (the control for temperature-diverse debates) --- isolating the effect of deliberation itself.

Model diversity assigns different base models to agents within one debate. We partition the pool into lineage classes (vendor $\times$ generation), forbid two same-class models per team, and stratify teams by empirical capability tier (strong/mid/weak, based on single-agent accuracy). We run experiments in four arms: (i) 80 three-model teams filling all ten tier compositions with balanced per-model appearances; (ii) two-model pairs in both majority compositions ($AAB$/$ABB$, $n{=}3$), which tests whether deliberation can overturn a built-in two-vote majority; (iii) four-model teams at $n{=}5$, duplicating once the strongest and once the weakest member; and (iv) mixed-ensemble controls replaying every team as single-round samples in matched proportions with majority voting, which serves as the model-diversity analogue of self-consistency while homogeneous debates at matching sizes serve as within-team controls.

\subsection{Diversity Metric}
\label{sec:divmetric}

To quantify persona diversity \emph{as perceived by the debating model itself}, we embed ``You are \{persona\}.''\ in the participant model's own space (mean-pooled, $\ell_2$-normalized final-layer states), form the cosine-similarity matrix $S$, and score a team by $D=\det(S)$, which corresponds to the Gram volume the personas span. Persona selection maximizes $D$ by exhaustive search over all tier-restricted combinations, per model, since persona semantics differ across families. We calculate $\log\det$ with diagonal jitter to guard underflow at large $n$. Further, quantized models embed on their unquantized parents. Our results are independent of the metric choice. Re-ranking every candidate trio under the max--min pairwise-distance objective reproduces the log-det ordering in all 23 models (Spearman $0.81$--$0.92$ over the full combination space), the minimum-determinant teams used in Section~\ref{sec:persona} fall at or below MaxMin's $1.1$th percentile, and every published MaxDet team lies above its $99.7$th (Appendix~\ref{app:metricrobust}). An additional verification with a dedicated sentence embedding model (all-MiniLM~\citep{wang2020minilm}) addresses the concern that anisotropic causal-LM states may measure style rather than semantics. We find that the ladder's direction remains unchanged (minimum-determinant teams fall at MiniLM's $7$th percentile, MaxDet teams at its $73$rd, median over models), while full-ranking correlations are deliberately modest (median $\rho\,{=}\,0.29$). Each model's own representation space contributes structure that a generic semantic embedder does not see, which is precisely the metric's design intent.

\section{Experimental Setup}

\subsection{Models}
We evaluate 23 open-weight models spanning eleven vendor families, three architecture types (dense transformer, mixture-of-experts, and hybrid state-space/attention), and a 180-fold range in parameter count (0.6B--109B total; Appendix~\ref{app:pool}, Table~\ref{tab:models}). Models above $\sim$27B run as published int4/FP8 quantizations and all others at native precision. The model-diversity teams (Section~\ref{sec:mixed}) draw from the 19 models available when team sampling was frozen; four later additions join all single-model conditions.

Biography evaluation uses a three-judge primary panel that never participates in debates --- Gemma-3-27B-it, GPT-OSS-20B, and GPT-OSS-120B --- with no judge sharing a model \emph{generation} with any participant: the pool's Gemma-4 members share a vendor with the Gemma-3 judge but not the model \emph{generation}, and the two OSS sizes deliberately separate inter-vendor agreement from judge-scale sensitivity. A fourth, supplementary judge (Meta Muse Glimmer-30B) deliberately \emph{shares} a vendor with two participants and is reported separately, to try and turn the judge--participant-coupling concern into a measured ablation rather than an assumption (Appendix~\ref{app:judges}).

\subsection{Tasks}
We evaluate on five benchmark tasks spanning mathematical reasoning, general and commonsense knowledge, as well as factual generation. The math task consists of 100 arithmetic expressions requiring order-of-operations reasoning. GSM8K~\citep{cobbe2021gsm8k} provides 100 grade-school math word problems. Both math tasks use exact numerical match with tolerance $10^{-4}$. \citet{hendrycks2021mmlu} contributes 100 four-choice academic questions, and \citet{talmor2019csqa} 100 five-choice commonsense questions (development split with seeded sampling without replacement). Multiple-choice answers are extracted position-sensitively and the explicit answer stated \emph{last} in a response wins. We make this choice because debate responses routinely quote other agents' answers before concluding and first-match extraction systematically mis-credits quoted answers. Problems for which no agent produces a parseable answer are scored incorrect rather than dropped.

The biography task requires generating factual biographies of 40 notable computer scientists, evaluated by a multi-judge, claim-based protocol: each generated biography is decomposed into atomic claims, and each judge classifies every claim against the reference article as supported, contradicted, or unverifiable. We report claim precision and, because reference articles are short, a complementary recall direction (reference facts checked against the generated biography), with inter-judge agreement (Cohen's $\kappa$) per direction and disagreement lists retained for adjudication. Verdicts are parsed by last-label occurrence, since reasoning judges deliberate before concluding, and ``unsupported'' is mapped to unverifiable so that substring matching cannot silently read it as supported.

Every condition answers identical problem sets (100 problems; biography 40 subjects; fixed sampling seed), so comparisons are paired by problem. Uncertainty is estimated by bootstrap~\citep{efron1979bootstrap} confidence intervals: paired by problem for debate-versus-control deltas, clustered over \emph{teams} where teams are the sampled unit, and correlations reported within model strata to avoid conflating composition with member capability. Serving infrastructure, generation settings, the full experimental grid, and the schematic of every matched pairing (Table~\ref{tab:pairings}) are in Appendix~\ref{app:infra}.

\section{Results}

\subsection{Main Results: Debate Helps Only as Much as Its Compute Budget}
\label{sec:main_results}

Table~\ref{tab:main} aggregates the four inline-scored tasks over the 23-model pool under the context-corrected policy of Section~\ref{sec:ctx}. MAD beats single-agent inference wherever a task has headroom ($+3.3$pp GSM8K, $+6.6$ MMLU, $+5.7$ CSQA; all $p<0.002$, Wilcoxon over models), with arithmetic at ceiling and unaffected. However, against the budget-matched control --- nine independent samples, same model and temperature spread, majority-voted --- the advantage disappears in all cases and it is indistinguishable from zero on math, GSM8K, and MMLU. On CSQA sampling \emph{wins} by $2.3$pp ($p=0.011$). At fixed budget, three rounds of deliberation buy nothing nine independent draws do not, while costing $1.6\times$ the wall-clock and $3.4\times$ the total tokens (generated tokens match at $1.06\times$; the gap can be attributed to debate re-reading its peers' transcripts each round; Appendix~\ref{app:cost}). In debate, later rounds re-process an ever-growing transcript while the control runs complete in one batched pass. Per-model details can be found in Appendix~\ref{app:permodel}.

\begin{table}[t]
\small
\centering
\caption{Mean accuracy (\%) per task over the 23-model pool (100 problems,
seed 42, context-corrected). Solo: one agent. Debate: $3\times3$. SC-9:
budget-matched self-consistency (9 samples, majority vote). Persona:
MaxDet persona debate. Bold: row best; $^*$: paired difference from Solo
has a 95\% bootstrap CI excluding zero. $^\dagger$Claim precision among
verifiable claims and vendor-balanced judge panel
(Section~\ref{sec:biography}); Mean averages the inline-scored tasks.}
\label{tab:main}
\label{tab:baselines}
\begin{tabular}{lcccc}
\toprule
Task & Solo & Debate & SC-9 & Persona \\
\midrule
Math & 91.2 & \textbf{91.5} & 90.6 & 90.0 \\
GSM8K & 87.9 & \textbf{91.2}$^*$ & 91.0$^*$ & 88.5 \\
MMLU & 69.9 & 76.5$^*$ & \textbf{76.8}$^*$ & 75.2$^*$ \\
CSQA & 72.6 & 78.3$^*$ & \textbf{80.6}$^*$ & 73.9 \\
Biography$^\dagger$ & 54.6 & 55.6 & \textbf{64.6}$^*$ & 51.8 \\
\midrule
Mean & 80.4 & 84.4 & \textbf{84.8} & 81.9 \\
\bottomrule
\end{tabular}

\end{table}

Furthermore, a rounds ablation over ten tier-spanning mixed teams makes the budget point mechanistic (Table~\ref{tab:rounds}) as essentially all of deliberation's benefit comes from the \emph{first} exchange (pooled $84.6\%$ at one round, $87.2$ at two, flat through five, CSQA declining past two). Transcripts confirm debate can repair errors: corrective flips outnumber harmful ones in all cases, and mixed teams overturn $63.7\%$ of wrong round-one majorities versus $42.9\%$ for homogeneous. Repairs concentrate in the first exchange, and voting over independent samples converts the same budget into accuracy at least as efficiently with no exchange at all.

\subsection{Personas Tax Small Models and Diversity Is Not the Culprit}
\label{sec:persona}

\begin{figure}[t]
\centering
\includegraphics[width=0.70\linewidth]{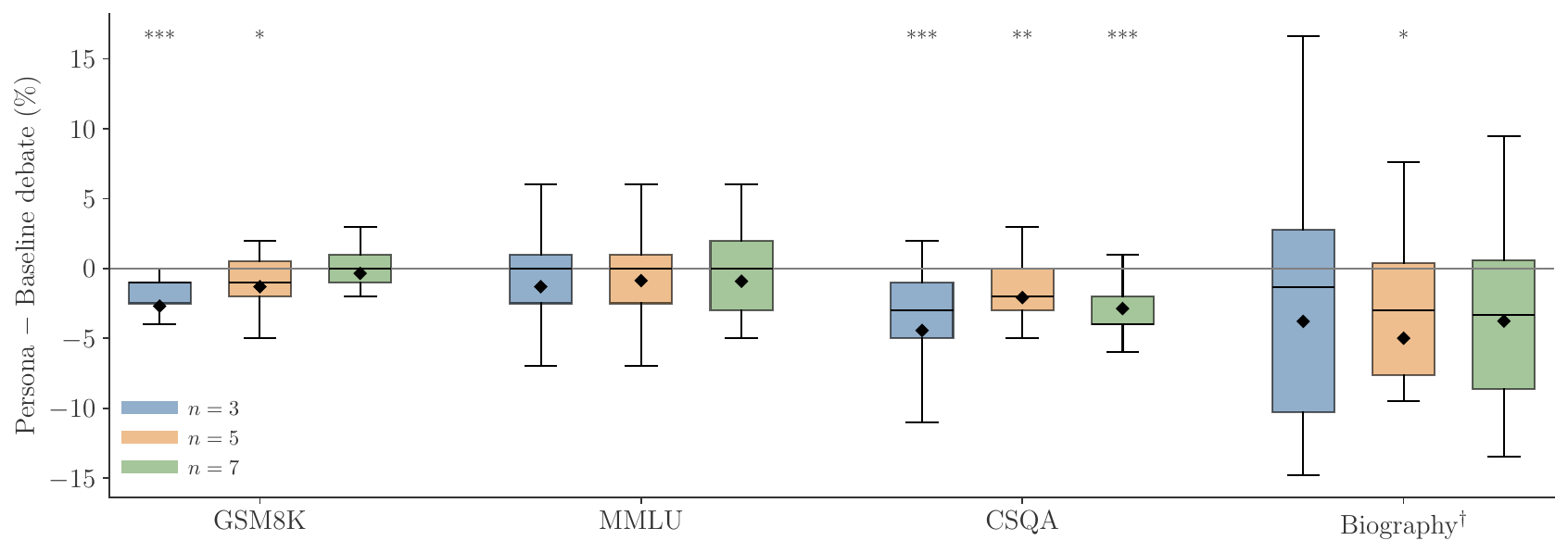}
\caption{Cost of maximally-diverse (MaxDet) persona teams by task and team
size: distribution over the 23-model pool of persona minus matched
baseline-debate accuracy (pp; seed 42, context-corrected). Diamonds:
means; stars: Wilcoxon vs.\ zero ($^{***}10^{-3}$, $^{**}10^{-2}$,
$^{*}0.05$). $^{\dagger}$Claim precision among verifiable claims under the
vendor-balanced panel of Section~\ref{sec:biography}.}
\label{fig:persona_by_task}
\label{fig:diversity}
\label{fig:diversity_by_task}
\end{figure}

If cognitive diversity drove debate gains, giving each agent a distinct reasoning persona should help, while giving the \emph{most} mutually distinct personas should help most. We test this hypothesis in two stages. First, for every model and team size, the persona team is the MaxDet-optimal combination, which was selected by exhaustive search over all $\binom{50}{n}$ combinations in that model's own embedding space (Section~\ref{sec:divmetric}). Figure~\ref{fig:persona_by_task} shows these maximally-diverse teams \emph{cost} accuracy relative to plain debate at the same model, team size, and budget. Results are significant on GSM8K at $n{=}3,5$ and on CSQA at every team size ($-2.1$ to $-4.4$pp, $p\le0.005$) and directionally negative on MMLU and on biography claim precision ($-3.8$ to $-5.0$pp, significant at $n{=}5$). Differences are pinned to zero on ceiling-bound arithmetic, which the figure therefore omits (49 of 69 paired cells differ by exactly zero; mean $-0.2$pp, $p=0.17$). A few per-model outliers (a Ministral collapse, a gemma-4-12B gain) are catalogued in Appendix~\ref{app:biography}.

The comparison above does not explain \emph{why} maximal teams lose: is diversity harmful, or is any persona harmful? The hypotheses make opposite predictions about less-diverse teams. We therefore re-enumerated all $\binom{50}{3}$ combinations per model (verifying the enumeration reproduces every published MaxDet team precisely) and built an 11-rung ladder: rung 0 being the \emph{minimum}-determinant team, rungs 1--9 the log-det decile maxima and rung 10 the MaxDet team. We then run all 230 new cells per task at $n{=}3$, $r{=}3$ under the context-corrected policy (zero empty final rounds fleet-wide).

\begin{figure}[t]
\centering
\includegraphics[width=0.65\linewidth]{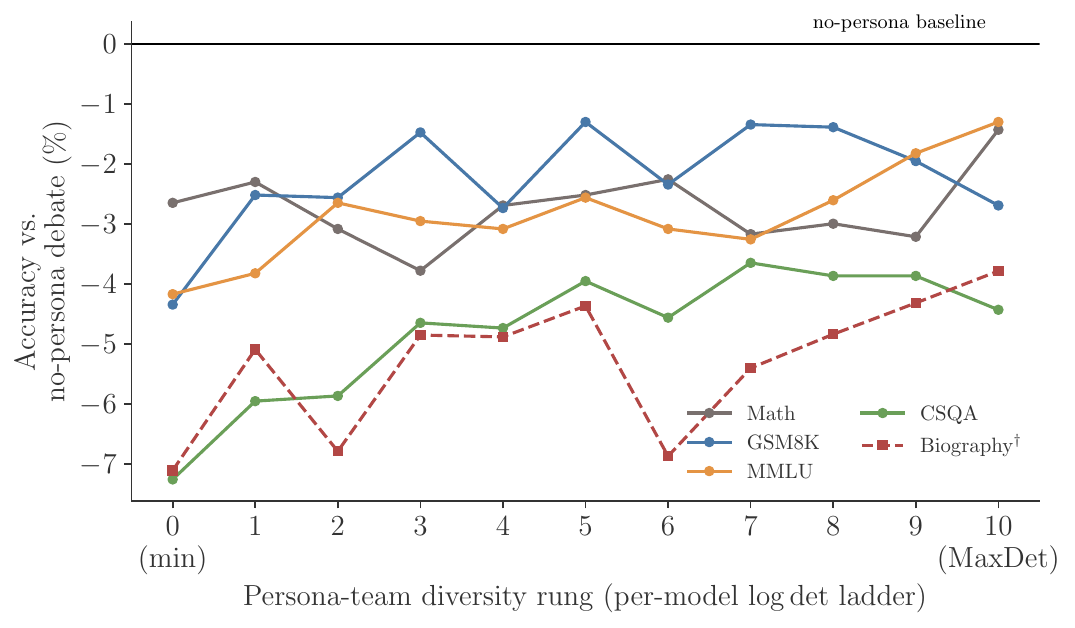}
\caption{Persona-diversity dose-response over the 23-model pool. Rung 0:
each model's minimum-determinant (most redundant) persona trio; rungs
1--9: log-det decile maxima; rung 10: the campaign MaxDet team. Lines:
mean delta vs.\ the no-persona baseline debate; bands: 95\% bootstrap over
models. $^\dagger$Biography (dashed) is claim precision under the
vendor-balanced judge panel.}
\label{fig:dose_response}
\end{figure}

Figure~\ref{fig:dose_response} shows the decomposition. The performance penalty is paid \emph{at the door}: the largest drop on every task is from baseline into rung 0; three near-clone personas cost $-4.3$pp on GSM8K ($p=0.012$), $-4.2$ on MMLU ($p=0.003$), and $-7.3$ on CSQA ($p=10^{-4}$), even before any diversity enters. From rung 0 to 10, holding ``has personas'' fixed, accuracy is flat to mildly \emph{increasing} ($+2.9$pp MMLU, $p=0.003$; within-model Spearman vs.\ rung positive on knowledge tasks). The pattern replicates on the judged task: biography claim precision is worst at rung 0 ($-7.1$pp, $p=0.048$), recovering $+3.3$pp by rung 10. The penalty in Figure~\ref{fig:persona_by_task} is therefore a \emph{persona} tax, not a \emph{diversity} tax: personas of any kind cost small models accuracy, and maximal diversity is the least harmful way to pay, and redundant personas the most. Notably, the dose does not act through observable disagreement. Within a model, rung leaves round-1 answer disagreement unchanged (mean within-stratum Spearman $-0.01$ across 92 model--task strata); the ladder manipulates the prompt frame and the outcome, not the vote distribution.

The cleanest test of parameter diversity holds everything else fixed: self-consistency in two flavors, all $N$ samples at base temperature versus spread over the debate arms' range ($0.5$--$1.0$). Paired over all 92 (model, task) cells, the spread changes nothing at any budget ($-0.11$ to $-0.41$pp, all $p\ge0.34$; no task moves more than $1.5$pp). The cheapest form of agent diversity neither helps nor hurts, meaning that the automatic temperature spread the debate arms receive cannot explain the debate-versus-ensemble comparisons below.

\subsection{Heterogeneous-Team Debate at a Matched Compute Budget}
\label{sec:mixed}

This subsection reports the model-diversity campaign: 230 heterogeneous teams evaluated on each of five tasks, of which the four automatically scorable tasks (math, GSM8K, MMLU, CommonsenseQA) are analysed here; judged biography results are reported in Section~\ref{sec:biography}.

\paragraph{Mixed-model debate does not beat matched-budget sampling.} The central comparison holds budget fixed: a three-model team debating three rounds versus an ensemble of nine single-round samples from the same models in the same proportions, majority-voted. Pairing each of the 80 core teams with its own control (Table~\ref{tab:mixed}), debate \emph{loses} by $1.04$pp (95\% CI $[-1.61,-0.52]$; 320 team--task cells), winning only 83 of 320 with a median delta of exactly zero. The deficit is significant on CSQA, math, and GSM8K, null on MMLU (Table~\ref{tab:mixed_bytask}), and replicates at a second problem seed ($-0.93$pp $[-1.45,-0.44]$). The other arms show the deficit is largest for the least diverse teams and vanishes for the most: strong-duplicated pairs lose $2.68$pp, while four-model teams at $n{=}5$ are indistinguishable from their fifteen-sample controls ($+0.04$/$+0.56$, null); the sole configuration where debate significantly \emph{beats} matched sampling is weak-duplicated quads on MMLU ($+2.80$pp). Weighting the ensembles' votes by member solo accuracy only widens their lead (Table~\ref{tab:weighted}; analysis in Appendix~\ref{app:mixed_bytask}). This sharpens rather than resolves the sampling critique of \citet{wynn2025}: mixing model families does not, by itself, buy an advantage over sampling more from the same mixture.

\begin{table}[t]
\small
\centering
\caption{Mixed-model debate against generation-budget-matched majority-vote
ensembles built from the same rosters (seed 42, 100 problems, four scorable
tasks). $\Delta$ = debate $-$ ensemble, 95\% CIs from a 10k bootstrap clustered over
team-configs, $^*$ marks intervals excluding zero. ``Strong/weak dup.''
duplicates the stronger or weaker member (pairs: AAB vs.\ ABB at $n{=}3$
vs.\ 9 samples; quads: $n{=}5$ vs.\ 15 samples).}
\label{tab:mixed}
\begin{tabular}{lcccc}
\toprule
Arm & Teams & Debate & Ensemble & $\Delta$ (pp) [95\% CI] \\
\midrule
Core triple ($n{=}3$) & 80 & 88.4 & 89.4 & $-1.04$ [$-1.61$, $-0.52$]$^*$ \\
Pair, strong dup.\ ($n{=}3$) & 15 & 83.0 & 85.7 & $-2.68$ [$-5.33$, $-0.65$]$^*$ \\
Pair, weak dup.\ ($n{=}3$) & 15 & 79.8 & 81.4 & $-1.63$ [$-4.62$, $+1.02$] \\
Quad, strong dup.\ ($n{=}5$) & 20 & 91.9 & 91.8 & $+0.04$ [$-0.39$, $+0.50$] \\
Quad, weak dup.\ ($n{=}5$) & 20 & 90.5 & 90.0 & $+0.56$ [$-0.34$, $+1.41$] \\
\bottomrule
\end{tabular}

\end{table}

\paragraph{Team accuracy tracks composition, not diversity.} Core-team accuracy is monotone in tier composition, falling from 93.7\% for all-strong teams (SSS) through 87.0\% (MMM) to 75.6\% for all-weak teams (WWW), with every intermediate composition ordered as its member tiers predict (Table~\ref{tab:tiers}). The same monotone ordering holds for the ensembles built from the same rosters. The ordering is explained by aggregate member capability rather than by heterogeneity: teams do not benefit from spanning more model families once member strength is accounted for, and the debate deficit concentrates in the mid-to-weak compositions.

\paragraph{The rescue effect belongs to the roster, not the debate.} Mixed teams do deliver one practically attractive result: on knowledge tasks, weak and mid-tier teams beat their own \emph{best member's} solo accuracy by $\ge$10 points in 19 team--task cells. But the matched ensembles reproduce and exceed it: they repair $+16.4$pp in those cells against debate's $+13.5$ and clear the 10-point rescue bar in 22 cells where debate does not (4 in the reverse direction). The recommendation this licenses concerns the roster: pooling votes from several different small models beats the best single model substantially, and skipping the debate makes it cheaper and slightly more accurate.

\paragraph{Caveats.} The debate arms receive a temperature spread while ensemble controls sample at base temperature; the temperature ablation closing Section~\ref{sec:persona} bounds this lever at $|0.4|$pp, far too small to account for the gaps above.

\subsection{Biography Generation}
\label{sec:biography}

The judged task tells the same story. Agents write biographies of 40 computer scientists; every claim is classified as supported, contradicted, or unverifiable against a reference article, and we score claim precision among verifiable claims under a vendor-balanced panel of three LLM judges that never participate in debates (Gemma-3-27B, GPT-OSS-20B, GPT-OSS-120B). Debate at $n{=}3$ edges solo generation ($55.6\%$ vs.\ $54.6\%$ claim precision) and persona teams reduce it to $51.8\%$ (Figure~\ref{fig:persona_by_task}). The budget-matched sampling control is again the strongest arm: nine independent biographies pooled through the same panel reach $64.6\%$ against debate's $55.6\%$ ($+8.9$pp on unrounded means). Seeing other agents' biographies \emph{hurts} precision relative to writing independently, extending the budget argument to the judged task. The persona penalty is significant only under the Gemma judge; judge vendor, not judge scale, dominates evaluation variance. Panel aggregation, the pooling mechanics behind the control's margin, refusal handling, per-judge splits, and per-model scores are in Appendix~\ref{app:biography}.

\subsection{Context-Window Ceilings}
\label{sec:ctx}

Multi-round debate has a failure mode single-round methods cannot have: the transcript grows every round, and long-form reasoning models emit tens of thousands of thinking tokens per turn. When the transcript approaches the \emph{serving} context window, later prompts leave no room to generate and agents return empty final answers, which scorers count as wrong. This destroyed cells for exactly five models, all long-form thinkers served at 16--32K windows; Qwen3.5-27B's seven-agent GSM8K, MMLU, and biography cells were 93--100\% empty in the final round. The failure masquerades as a debate result: the temperature-matched debate-versus-sampling comparison stood at $-1.84$pp as run, $-0.30$ excluding bound cells, and $-0.06$ after re-running every affected cell at 131K (276 paired cells). The entire apparent deficit beyond noise was window overflow, not deliberation failure (one cell: Qwen3.5-27B on GPQA, $0.24$ at 16K vs.\ $0.82$ at 131K). Figure~\ref{fig:window_curve} shows accuracy recovering monotonically exactly as the empty-response fraction falls to zero. We treat window size as a measurement correction (all headline numbers use the corrected runs). The deployment rule is that the window must accommodate agents $\times$ rounds $\times$ per-turn budget, or debate degrades silently and benchmarks blame the method.

\section{Discussion}

Before the dose-response experiment, the natural reading of Figure~\ref{fig:persona_by_task} was that \emph{extreme diversity} breaks debate: agents communicating in koans cannot build on each other's arguments, so maximally-distinct personas cause consensus failure. The ladder disconfirms this as the driver: the most \emph{redundant} persona teams (three near-clone single-metric evaluators) hurt more than the koan-speakers on every task, and increasing diversity weakly improves accuracy. Communication breakdown between mutually alien personas cannot explain a penalty that is largest when the personas are nearly identical.

What survives is a sharpened \emph{capacity-tax} account: adopting any assigned cognitive frame consumes representational resources small models need for the task, roughly independently of which frame. The tax is largest on knowledge retrieval (CSQA $-7.3$pp at rung 0), smallest at the arithmetic ceiling, and shrinks with model capability (Spearman $+0.46$, $p=0.027$): stronger models afford the costume. Nor is it mere prompt overhead: persona descriptions span only 11--19 words while per-persona drag spans 21pp, and description length explains under $9\%$ of between-persona variance ($R^2{=}0.087$ on word count, and character count is uncorrelated outright, $r{=}-0.01$; 1{,}605 paired agent-slot observations). The cost tracks what the persona says, not how long it is.

Why does diversity then help slightly, conditional on paying? Redundant personas add \emph{correlated} bias that majority voting cannot remove; diverse personas at least decorrelate the damage, which is the ensemble-theoretic prediction, visible only once the tax is separated out. Nor is the harm a small-team voting artifact: it persists, and on knowledge tasks deepens, at five and seven agents. A static signature of the tax may live in the weights themselves: models that represent the persona set in a lower-volume region of their own embedding space pay a larger penalty ($\rho\approx-0.4$), though at 23 models this is confounded with capability and we do not lean on it here; whether prompting costs are predictable from geometry alone is a question we develop in separate work.

The capacity-tax account also explains an asymmetry across diversity types: model diversity is harmless where persona diversity is harmful. Different models bring genuinely different capabilities without imposing load on any single agent: each operates in its natural mode, and mixed teams suffer no analogue of the persona tax. But harmless is not helpful. The matched-ensemble controls show that everything model diversity contributes (error decorrelation across families, complementary failure modes) is captured by \emph{voting} over the mixed roster; the deliberation layered on top subtracts a little (Table~\ref{tab:mixed}). If personas fail because small models cannot afford the costume, and model diversity's benefits are realized by aggregation alone, then no form of agent diversity we tested gives debate an edge over sampling; whether larger models can convert diversity into genuinely deliberative gains is the natural follow-up.

\section{Limitations and Future Work}

Several limitations constrain the scope of our conclusions. Our findings are established for small open-weight models; the capacity account of the persona penalty predicts that larger models could adopt unconventional cognitive frames without paying the tax; the pool's largest model already hints at it: Llama-4-Scout's maximally-diverse persona teams beat its matched sampling control on MMLU ($87.0$ vs.\ $83.0$) and CSQA ($91.0$ vs.\ $89.0$), though one model is an anecdote, not a trend, and the budget accounting could differ where a single frontier model dwarfs the cost of extra samples. Seed sensitivity is addressed but not eliminated: the headline mixed-model deficit replicates at a second problem seed ($-0.93$ vs.\ $-1.04$pp), and the homogeneous grid's second-seed slice correlates $0.978$ with the first, but per-task significance patterns do shift between seeds and each cell rests on 100 problems (40 biography subjects).

Our persona intervention is prompt-based, and although the dose-response ladder spans each model's full determinant range, all rungs are determinant-\emph{selected}; task-aligned persona choice, fine-tuned styles, or tool and information asymmetries could produce cognitive diversity that prompting cannot. Structured single-agent test-time methods (tree-of-thoughts search~\citep{yao2023tree}, iterative self-refinement~\citep{madaan2023selfrefine}) and objectively verifiable domains such as BIG-Bench Hard or code generation remain untested here; they are natural next mechanisms to hold to the same budget-matched bar. The biography conclusions inherit the judges' disagreement structure: we expose it (vendor-balanced scoring, per-judge splits, $\kappa$ matrices) rather than resolve it, absent human claim-level annotation. Our tasks emphasize verifiable answers; domains valuing novelty or synthesis may reward diversity differently. Future work should learn team composition rather than fix it: the pool already contains matched pairs for distillation provenance (R1-Distill-32B vs.\ Qwen3-32B, identically quantized) and MoE-versus-dense at one rung (Qwen3.5-35B-A3B vs.\ 27B), but integrating them into the stratified team design requires resampling the team space.

\section{Conclusion}

Across 23 open-weight models, five tasks, three diversity axes, and 5,500+ runs with matched sampling controls, the cognitive-diversity hypothesis fails on every axis. Persona prompting reduces accuracy, and the dose-response ladder shows the loss is a persona tax rather than a diversity effect. Temperature diversity changes nothing. Heterogeneous teams realize all of their benefits through majority voting alone, a result that replicates across problem seeds. Debate's genuine gains over single-agent inference are an ensemble-sampling effect: nearly all benefit arrives with the first exchange, and none survives budget matching, at $1.6\times$ the wall-clock and $3.4\times$ the token cost.

Multi-round debate also carries a silent measurement hazard: transcript overflow, large enough to have manufactured our entire apparent debate deficit before correction. The one boundary case closed the same way. On the hardest slice, debate's directional edge over sampling came from a generation-budget cap handicapping the control, and correcting it leaves that comparison null too (Appendix~\ref{app:p05}). The transcripts locate the inefficiency precisely: debate \emph{is} net-corrective (helpful flips outnumber harmful ones roughly two to one, $7.6\%$ vs.\ $4.0\%$ of agent--problem pairs), yet voting over independent samples converts the same generation budget into at least as much accuracy with no exchange at all. The language-mediated channel is not destructive, merely expensive.

Iteration pays \emph{inside} a model's forward process and stops paying when routed through exchanged text between agents: the strongest single lever in our grid is long-form serial reasoning, aggregated in parallel by a vote. That is the design point of the depth-recurrence lineage, from parameter-tied iteration~\citep{dehghani2019universal} to latent test-time reasoning and adaptive recursive depth~\citep{geiping2025scaling,saunshi2025reasoning,bae2025mixture}: keep the iteration, drop the transcript. We release the full budget-matched, contamination-checked grid as the baseline that any proposed debate mechanism should be required to beat.

\newpage

\subsubsection*{Reproducibility Statement}
All models are open-weight releases served on a single engine version;
every experimental family, its role, run count, and problem-sampling seed
are enumerated in Appendix~\ref{app:design}. Every number in the paper is
produced by a deterministic script: figure and table emitters (with fixed
bootstrap seeds) live in \texttt{paper/figures/} of the accompanying code
release, scoring uses one shared scorer, and per-file scored outputs,
judge-verdict tables, persona ladders with full determinant distributions,
and the fleet manifest are released as CSVs. Persona-team selection is
exactly reproducible: re-enumeration of each model's combinatorial persona
space reproduces every published team, a check asserted programmatically
in the released pipeline.

\ifdefined\arxivbuild
\subsubsection*{Acknowledgments}
We thank the Boston Children's Hospital Research Computing group for computational resources and support on the E3 cluster, and Vennela Jonnala for contributions to an earlier phase of this project. This work was conducted with grant support from the Research Innovation in Support of Excellence (RISE) Award at Cincinnati Children's Hospital.
\fi

\subsubsection*{AI Use Statement}
Beyond conventional writing assistance, AI systems were used as research
infrastructure throughout this project, under continuous human direction.
A large language model agent (Claude, Anthropic) reviewed the analysis code
and performed campaign orchestration and scheduling across two
compute environments, including failure diagnosis and recovery, data aggregation and
verification. All research questions, experimental designs, scope rulings, and interpretive claims are
the authors'. No AI system is an
author of this work, consistent with venue policy; this statement is our
account of what the tools did.

{
\small

\bibliographystyle{iclr2027_conference}
\bibliography{references}
}

\appendix

\section{The Model Pool}
\label{app:pool}

\begin{table}[H]
\footnotesize
\centering
\caption{The 23-model participant pool. Class = vendor$\times$generation lineage used for the model-diversity constraint. Quantized checkpoints are official releases. The four models below the rule joined after model-diversity team sampling was frozen and appear in all single-model conditions only.}
\label{tab:models}
\begin{tabular}{llll}
\toprule
Model & Size & Class & Notes \\
\midrule
Qwen3-0.6B & 0.6B & Qwen3 & \\
LFM2.5-1.2B & 1.2B & LFM2.5 & hybrid conv/attn \\
VibeThinker-1.5B & 1.5B & VibeThinker & reasoning FT \\
Llama-3.2-3B-Instruct & 3B & Llama-3.2 & \\
Phi-4-mini & 3.8B & Phi-4 & synthetic-data \\
Nemotron-3-Nano-4B & 4B & Nemotron-3 & Mamba hybrid \\
Mistral-7B-Instruct-v0.3 & 7B & Mistral-7B & \\
Olmo-3-7B-Instruct & 7B & Olmo-3 & fully open \\
Ministral-3-8B & 8B & Ministral-3 & \\
Granite-4.1-8B & 8B & Granite-4 & \\
LFM2.5-8B-A1B & 8B & LFM2.5 & MoE, 1B active \\
Qwen3.5-9B & 9B & Qwen3.5 & \\
Gemma-4-12B-it & 12B & Gemma-4 & \\
Qwen3-14B & 14B & Qwen3 & \\
Phi-4 & 14B & Phi-4 & synthetic-data \\
Qwen3.5-27B & 27B & Qwen3.5 & GPTQ-Int4 \\
Nemotron-3-Nano-30B-A3B & 30B & Nemotron-3 & MoE, FP8 \\
Qwen3-32B & 32B & Qwen3 & W4A16 \\
Llama-4-Scout & 109B & Llama-4 & MoE 17B act., W4A16 \\
\midrule
R1-Distill-Qwen-14B & 14B & DeepSeek-R1 & distilled \\
Gemma-4-26B-A4B & 26B & Gemma-4 & MoE, FP8 \\
R1-Distill-Qwen-32B & 32B & DeepSeek-R1 & distilled, W4A16 \\
Qwen3.5-35B-A3B & 35B & Qwen3.5 & MoE, GPTQ-Int4 \\
\bottomrule
\end{tabular}
\end{table}

\section{Experimental Design and Infrastructure}
\label{app:design}
\label{app:infra}

Table~\ref{tab:design} maps every experiment family run to
its role in the argument: \emph{core} families instantiate the paper's
questions; \emph{control} families are the generation-budget-matched
comparison arms; \emph{ablation} families probe robustness (seeds, rounds,
benchmark variants); the context-window family is a \emph{correction}, designed as 
a measurement fix whose necessity is itself a finding
(Section~\ref{sec:ctx}) and not merely an ablation. ``Runs'' are debate/sampling
executions of 100 problems each (biography: 40 subjects); the judging row
counts judge$\times$file evaluation passes rather than generation runs.

\begin{table}[H]
\scriptsize
\centering
\caption{Every experiment family, its role, model pool, task coverage, run
count, and problem-sampling seed. The 23-model pool comprises 16 local
workstation models, 3 cluster-only models, and 4 later-added models; the
mixed-model families draw teams from the 19 models available at team-freeze
time.}
\label{tab:design}
\begin{tabular}{llllrl}
\toprule
Family & Role & Pool & Tasks & Runs & Seed \\
\midrule
Homogeneous baseline (solo + debate $n{\in}\{3,5,7\}$) & core & 23 & 5 & 460 & 42 \\
MaxDet persona debate ($n{\in}\{3,5,7\}$) & core & 23 & 5 & 345 & 42 \\
Self-consistency, budget-matched ($N{\in}\{9,15,21\}$, uniform + spread) & control & 23 & 5 & 690 & 42 \\
Seed-43 homogeneous slice (solo + debate $n{=}3$) & ablation & 19 & 5 & 190 & 43 \\
\midrule
Mixed-model core triples ($n{=}3$, 80 teams) & core & 19 & 5 & 400 & 42 \\
Mixed-model pairs (AAB/ABB, 15 pairs) & core & 19 & 5 & 150 & 42 \\
Mixed-model quads ($n{=}5$, dup-strong/weak, 20 teams) & core & 19 & 5 & 200 & 42 \\
Matched mixed ensembles (core: $9{\times}1$) & control & 19 & 5 & 400 & 42 \\
Matched mixed ensembles (pair $9{\times}1$ / quad $15{\times}1$) & control & 19 & 5 & 350 & 42 \\
Mixed rounds ablation ($r{\in}\{1,2,4,5\}$, 10 teams) & ablation & 19 & 5 & 200 & 42 \\
Seed-43 mixed replication (core + matched ensemble) & ablation & 19 & 5 & 800 & 43 \\
\midrule
Context-window grid + ctx-max re-runs (5 bound models) & correction & 5 & 5 & 170 & 42 \\
Persona dose-response ladder (rungs 0--9, $n{=}3$) & core & 23 & 5 & 1{,}150 & 42 \\
MMLU-Pro + GSM-Symbolic slices (solo/debate/SC-9) & ablation & 23 & 2 & 138 & 42 \\
GPQA Diamond (full pool + budget-liberated arm) & ablation & 23 & 1 & 93 & 42 \\
\midrule
Biography judging: 4-judge panel over all judged biographies & eval & 4 judges & 1 & 2{,}720 & --- \\
\bottomrule
\end{tabular}

\end{table}

All experiments run on the same engine version -- vLLM 0.25.1~\citep{kwon2023vllm} -- across a dual RTX 3090 workstation (to demonstrate accessibility) and an institutional SLURM HPC cluster with multiple L40 and A100 nodes. Debate is batched round-major (one engine call per agent per round across all problems). Generation uses \texttt{top\_p=1.0} and \texttt{temperature=1.0} unless a diversity condition varies it. Context windows are per-model and engine-servable maxima. When an accumulating transcript exceeds a window the response returns empty and scores incorrect, and we report overflow rates per condition (Section~\ref{sec:ctx}).

The grid comprises multiple configurations: baselines at $n\in\{1,3,5,7\}$ and MaxDet persona teams at $n\in\{3,5,7\}$ for every model and task; budget-matched self-consistency at $N\in\{9,15,21\}$ in both temperature flavors; the 1{,}150-run model-diversity campaign (stratified triples, pair compositions, duplication-rule quads, and matched roster-ensembles); a 1{,}150-run persona dose-response fleet (Section~\ref{sec:persona}); and ablations for problem-seed replication, pair/quad matched controls, and debate rounds $r\in\{1,2,4,5\}$. Table~\ref{tab:pairings} is the schematic of every matched pairing and the temperature policy on each side. Appendix~\ref{app:design} maps every family to its role (core, control, ablation, correction) with run counts.

\begin{table}[H]
\centering
\caption{
Matched-control pairings. Every debate arm is compared against a control
built from the same model roster at the same generation budget
(generations per problem) in ``two flavors'': uniform temperature
($T{=}1.0$) and the same per-agent spread that the debate uses, which isolates
deliberation from sampling stochasticity. Persona debates fix $T{=}1.0$ so
the persona effect is not confounded with temperature and their like-for-like
control is uniform-temperature self-consistency. T, temperature; SC, self-consistency.}
\label{tab:pairings}
\small
\begin{tabular}{@{}llcll@{}}
\toprule
Debate arm & Temp.\ policy & Budget & Matched control & Temp.\ policy \\
\midrule
Homog.\ $n{=}3$, $r{=}3$ & spread $0.5$--$1.0$ & 9 & SC-9 (two flavors) & $T{=}1.0$ / spread \\
Homog.\ $n{=}5$, $r{=}3$ & spread $0.5$--$1.0$ & 15 & SC-15 (two flavors) & $T{=}1.0$ / spread \\
Homog.\ $n{=}7$, $r{=}3$ & spread $0.5$--$1.0$ & 21 & SC-21 (two flavors) & $T{=}1.0$ / spread \\
Persona $n{\in}\{3,5,7\}$, $r{=}3$ & $T{=}1.0$ fixed & $3n$ & SC-$3n$ uniform & $T{=}1.0$ \\
Persona rungs 0--10, $n{=}3$ & $T{=}1.0$ fixed & 9 & no-persona debate & spread $0.5$--$1.0$ \\
Mixed triple $1{+}1{+}1$, $r{=}3$ & spread $0.5$--$1.0$ & 9 & roster ens.\ $3{\times}3$, $r{=}1$ & $T{=}1.0$ \\
Mixed pair AAB/ABB, $r{=}3$ & spread $0.5$--$1.0$ & 9 & roster ens.\ $9{\times}1$, $r{=}1$ & $T{=}1.0$ \\
Mixed quad $n{=}5$ (dup-s/w), $r{=}3$ & spread $0.5$--$1.0$ & 15 & roster ens.\ $15{\times}1$, $r{=}1$ & $T{=}1.0$ \\
Rounds $r{\in}\{1,2,4,5\}$, $n{=}3$ & spread $0.5$--$1.0$ & $3r$ & $r{=}3$ arm (budget varies) & spread $0.5$--$1.0$ \\
\bottomrule
\end{tabular}

\end{table}

\section{Per-Model Results}
\label{app:permodel}

Tables~\ref{tab:permodel_math}--\ref{tab:permodel_csqa} report the
per-model accuracies underlying Table~\ref{tab:main}: each of the 23 pool
models under Solo, Debate ($3\times3$), matched SC-9, and
maximally-diverse persona debate, at seed 42 under the ctx-max policy.

\begin{table}[H]\small\centering
\caption{Per-model accuracy (\%), Math.}
\label{tab:permodel_math}
\begin{tabular}{lcccc}
\toprule
Model & Solo & Debate & SC-9 & Persona \\
\midrule
LFM2.5-1.2B & \textbf{100.0} & \textbf{100.0} & 99.0 & 99.0 \\
LFM2.5-8B-A1B & \textbf{100.0} & \textbf{100.0} & \textbf{100.0} & 97.0 \\
Llama-3.2-3B & 35.0 & 37.0 & 29.0 & \textbf{39.0} \\
Llama-4-Scout-17B-16E (W4A16) & \textbf{100.0} & \textbf{100.0} & \textbf{100.0} & \textbf{100.0} \\
Ministral-3-8B & 98.0 & 96.0 & \textbf{99.0} & 54.0 \\
Mistral-7B-v0.3 & 7.0 & 9.0 & \textbf{18.0} & 9.0 \\
NVIDIA-Nemotron-3-Nano-30B-A3B-FP8 & 98.0 & \textbf{100.0} & \textbf{100.0} & 97.0 \\
NVIDIA-Nemotron-3-Nano-4B-BF16 & 97.0 & \textbf{100.0} & \textbf{100.0} & \textbf{100.0} \\
Olmo-3-7B & \textbf{100.0} & \textbf{100.0} & \textbf{100.0} & \textbf{100.0} \\
Phi-4-mini-instruct & \textbf{96.0} & \textbf{96.0} & 67.0 & 87.0 \\
Qwen3-0.6B & 97.0 & \textbf{100.0} & \textbf{100.0} & 99.0 \\
Qwen3-14B & \textbf{100.0} & \textbf{100.0} & \textbf{100.0} & \textbf{100.0} \\
Qwen3-32B (W4A16) & \textbf{100.0} & \textbf{100.0} & \textbf{100.0} & \textbf{100.0} \\
Qwen3.5-27B (GPTQ) & \textbf{100.0} & \textbf{100.0} & \textbf{100.0} & \textbf{100.0} \\
Qwen3.5-35B-A3B (GPTQ) & \textbf{100.0} & \textbf{100.0} & \textbf{100.0} & \textbf{100.0} \\
Qwen3.5-9B & \textbf{100.0} & \textbf{100.0} & \textbf{100.0} & 99.0 \\
R1-Distill-Qwen-14B & 99.0 & \textbf{100.0} & \textbf{100.0} & \textbf{100.0} \\
R1-Distill-Qwen-32B-w4a16 & 99.0 & \textbf{100.0} & \textbf{100.0} & 98.0 \\
VibeThinker-1.5B & \textbf{100.0} & \textbf{100.0} & \textbf{100.0} & \textbf{100.0} \\
gemma-4-12B-it & \textbf{100.0} & \textbf{100.0} & \textbf{100.0} & \textbf{100.0} \\
gemma-4-26B-A4B-it-FP8-dy & \textbf{100.0} & \textbf{100.0} & \textbf{100.0} & \textbf{100.0} \\
granite-4.1-8b & \textbf{100.0} & \textbf{100.0} & \textbf{100.0} & \textbf{100.0} \\
phi-4 & 72.0 & 66.0 & 72.0 & \textbf{93.0} \\
\bottomrule
\end{tabular}

\end{table}

\begin{table}[H]\small\centering
\caption{Per-model accuracy (\%), GSM8K.}
\label{tab:permodel_gsm}
\begin{tabular}{lcccc}
\toprule
Model & Solo & Debate & SC-9 & Persona \\
\midrule
LFM2.5-1.2B & 70.0 & 74.0 & \textbf{75.0} & 70.0 \\
LFM2.5-8B-A1B & 89.0 & 96.0 & \textbf{99.0} & 89.0 \\
Llama-3.2-3B & 46.0 & 78.0 & \textbf{85.0} & 60.0 \\
Llama-4-Scout-17B-16E (W4A16) & \textbf{96.0} & 95.0 & \textbf{96.0} & 94.0 \\
Ministral-3-8B & 91.0 & 94.0 & \textbf{95.0} & 88.0 \\
Mistral-7B-v0.3 & 43.0 & \textbf{55.0} & 40.0 & 43.0 \\
NVIDIA-Nemotron-3-Nano-30B-A3B-FP8 & 95.0 & \textbf{96.0} & 95.0 & 95.0 \\
NVIDIA-Nemotron-3-Nano-4B-BF16 & 94.0 & \textbf{95.0} & \textbf{95.0} & 94.0 \\
Olmo-3-7B & 92.0 & 93.0 & \textbf{94.0} & 91.0 \\
Phi-4-mini-instruct & 84.0 & 88.0 & \textbf{92.0} & 86.0 \\
Qwen3-0.6B & 78.0 & 79.0 & \textbf{80.0} & 78.0 \\
Qwen3-14B & \textbf{97.0} & 96.0 & 96.0 & 96.0 \\
Qwen3-32B (W4A16) & 95.0 & 95.0 & \textbf{96.0} & 94.0 \\
Qwen3.5-27B (GPTQ) & 96.0 & \textbf{98.0} & 93.0 & \textbf{98.0} \\
Qwen3.5-35B-A3B (GPTQ) & 96.0 & \textbf{98.0} & 94.0 & \textbf{98.0} \\
Qwen3.5-9B & 93.0 & \textbf{97.0} & 90.0 & 94.0 \\
R1-Distill-Qwen-14B & 96.0 & 98.0 & \textbf{99.0} & 97.0 \\
R1-Distill-Qwen-32B-w4a16 & 95.0 & 96.0 & \textbf{98.0} & 95.0 \\
VibeThinker-1.5B & 94.0 & 94.0 & \textbf{95.0} & 93.0 \\
gemma-4-12B-it & \textbf{97.0} & \textbf{97.0} & \textbf{97.0} & 96.0 \\
gemma-4-26B-A4B-it-FP8-dy & \textbf{98.0} & 97.0 & \textbf{98.0} & 96.0 \\
granite-4.1-8b & 93.0 & 94.0 & 94.0 & \textbf{96.0} \\
phi-4 & 94.0 & 94.0 & \textbf{98.0} & 94.0 \\
\bottomrule
\end{tabular}

\end{table}

\begin{table}[H]\small\centering
\caption{Per-model accuracy (\%), MMLU.}
\label{tab:permodel_mmlu}
\begin{tabular}{lcccc}
\toprule
Model & Solo & Debate & SC-9 & Persona \\
\midrule
LFM2.5-1.2B & 45.0 & 55.0 & \textbf{57.0} & \textbf{57.0} \\
LFM2.5-8B-A1B & 61.0 & \textbf{80.0} & 79.0 & 74.0 \\
Llama-3.2-3B & 17.0 & \textbf{49.0} & 45.0 & 42.0 \\
Llama-4-Scout-17B-16E (W4A16) & 83.0 & 84.0 & 83.0 & \textbf{87.0} \\
Ministral-3-8B & 64.0 & \textbf{79.0} & 78.0 & 68.0 \\
Mistral-7B-v0.3 & 40.0 & \textbf{60.0} & 59.0 & 45.0 \\
NVIDIA-Nemotron-3-Nano-30B-A3B-FP8 & 82.0 & \textbf{86.0} & 82.0 & 82.0 \\
NVIDIA-Nemotron-3-Nano-4B-BF16 & 73.0 & 75.0 & \textbf{76.0} & 75.0 \\
Olmo-3-7B & 66.0 & 69.0 & 69.0 & \textbf{72.0} \\
Phi-4-mini-instruct & 52.0 & 58.0 & 61.0 & \textbf{64.0} \\
Qwen3-0.6B & 54.0 & 56.0 & \textbf{60.0} & 54.0 \\
Qwen3-14B & 84.0 & \textbf{87.0} & \textbf{87.0} & 84.0 \\
Qwen3-32B (W4A16) & 86.0 & \textbf{89.0} & 88.0 & \textbf{89.0} \\
Qwen3.5-27B (GPTQ) & 87.0 & 91.0 & 90.0 & \textbf{92.0} \\
Qwen3.5-35B-A3B (GPTQ) & \textbf{92.0} & 89.0 & 87.0 & 88.0 \\
Qwen3.5-9B & 85.0 & \textbf{89.0} & 87.0 & 88.0 \\
R1-Distill-Qwen-14B & 75.0 & 82.0 & \textbf{85.0} & 82.0 \\
R1-Distill-Qwen-32B-w4a16 & 78.0 & \textbf{86.0} & \textbf{86.0} & \textbf{86.0} \\
VibeThinker-1.5B & 58.0 & 59.0 & \textbf{69.0} & 60.0 \\
gemma-4-12B-it & 83.0 & 85.0 & 83.0 & \textbf{86.0} \\
gemma-4-26B-A4B-it-FP8-dy & 87.0 & 87.0 & 87.0 & \textbf{89.0} \\
granite-4.1-8b & 67.0 & 77.0 & \textbf{82.0} & 77.0 \\
phi-4 & \textbf{88.0} & 87.0 & 86.0 & \textbf{88.0} \\
\bottomrule
\end{tabular}

\end{table}

\begin{table}[H]\small\centering
\caption{Per-model accuracy (\%), CSQA.}
\label{tab:permodel_csqa}
\begin{tabular}{lcccc}
\toprule
Model & Solo & Debate & SC-9 & Persona \\
\midrule
LFM2.5-1.2B & 55.0 & 62.0 & \textbf{64.0} & 61.0 \\
LFM2.5-8B-A1B & 71.0 & 80.0 & \textbf{81.0} & 75.0 \\
Llama-3.2-3B & 26.0 & 60.0 & \textbf{69.0} & 53.0 \\
Llama-4-Scout-17B-16E (W4A16) & 85.0 & 89.0 & 89.0 & \textbf{91.0} \\
Ministral-3-8B & 61.0 & 72.0 & \textbf{76.0} & 39.0 \\
Mistral-7B-v0.3 & 49.0 & 69.0 & \textbf{73.0} & 62.0 \\
NVIDIA-Nemotron-3-Nano-30B-A3B-FP8 & 86.0 & 87.0 & \textbf{88.0} & 82.0 \\
NVIDIA-Nemotron-3-Nano-4B-BF16 & 73.0 & \textbf{81.0} & 79.0 & 76.0 \\
Olmo-3-7B & \textbf{83.0} & \textbf{83.0} & \textbf{83.0} & 72.0 \\
Phi-4-mini-instruct & 60.0 & 66.0 & \textbf{69.0} & 65.0 \\
Qwen3-0.6B & 47.0 & 50.0 & \textbf{60.0} & 50.0 \\
Qwen3-14B & \textbf{85.0} & 84.0 & \textbf{85.0} & 82.0 \\
Qwen3-32B (W4A16) & \textbf{88.0} & 85.0 & 86.0 & 84.0 \\
Qwen3.5-27B (GPTQ) & 89.0 & 89.0 & \textbf{91.0} & 90.0 \\
Qwen3.5-35B-A3B (GPTQ) & 88.0 & \textbf{90.0} & 85.0 & 87.0 \\
Qwen3.5-9B & 83.0 & \textbf{88.0} & 85.0 & 84.0 \\
R1-Distill-Qwen-14B & 76.0 & 84.0 & \textbf{89.0} & 80.0 \\
R1-Distill-Qwen-32B-w4a16 & 80.0 & 84.0 & \textbf{90.0} & 81.0 \\
VibeThinker-1.5B & 57.0 & 65.0 & \textbf{74.0} & 63.0 \\
gemma-4-12B-it & \textbf{83.0} & 80.0 & 81.0 & 82.0 \\
gemma-4-26B-A4B-it-FP8-dy & \textbf{90.0} & 88.0 & 88.0 & 83.0 \\
granite-4.1-8b & 70.0 & \textbf{81.0} & \textbf{81.0} & 80.0 \\
phi-4 & 85.0 & 85.0 & \textbf{88.0} & 78.0 \\
\bottomrule
\end{tabular}

\end{table}

\section{Mixed-Model Arms by Task}
\label{app:mixed_bytask}

Table~\ref{tab:mixed_bytask} decomposes the pooled quad null of Table~\ref{tab:mixed} by task: debate is negative where significant on math, GSM8K, and CSQA, but the weak-duplicated quad arm on MMLU is the one configuration where debate significantly \emph{beats} matched sampling ($+2.80$ $[+0.90, +4.60]$).

\begin{table}[H]\small\centering
\caption{Per-task debate-minus-matched-ensemble deltas (pp) for every
mixed-model arm of Table~\ref{tab:mixed} (95\% CIs, bootstrap clustered
over team-configs; $^*$ excludes zero).}
\label{tab:mixed_bytask}
\begin{tabular}{lcccc}
\toprule
Arm & Math & GSM8K & MMLU & CSQA \\
\midrule
Core triple ($n{=}3$) & $-1.45$ {\scriptsize$[-2.6,-0.4]$}$^*$ & $-1.10$ {\scriptsize$[-1.6,-0.6]$}$^*$ & $-0.01$ {\scriptsize$[-0.9,+0.8]$} & $-1.61$ {\scriptsize$[-2.6,-0.7]$}$^*$ \\
Pair, strong dup.\ ($n{=}3$) & $-3.53$ {\scriptsize$[-8.1,+0.1]$} & $-1.73$ {\scriptsize$[-4.1,+0.3]$} & $-3.27$ {\scriptsize$[-6.2,-0.7]$}$^*$ & $-2.20$ {\scriptsize$[-4.7,+0.3]$} \\
Pair, weak dup.\ ($n{=}3$) & $-0.53$ {\scriptsize$[-6.5,+5.8]$} & $-1.67$ {\scriptsize$[-3.9,+0.5]$} & $-2.13$ {\scriptsize$[-5.9,+1.3]$} & $-2.20$ {\scriptsize$[-5.4,+0.9]$} \\
Quad, strong dup.\ ($n{=}5$) & $-0.10$ {\scriptsize$[-0.3,+0.0]$} & $-0.60$ {\scriptsize$[-1.2,+0.1]$} & $+0.95$ {\scriptsize$[-0.4,+2.3]$} & $-0.10$ {\scriptsize$[-0.9,+0.7]$} \\
Quad, weak dup.\ ($n{=}5$) & $+0.15$ {\scriptsize$[-0.4,+0.9]$} & $-1.25$ {\scriptsize$[-2.2,-0.5]$}$^*$ & $+2.80$ {\scriptsize$[+0.9,+4.6]$}$^*$ & $+0.55$ {\scriptsize$[-1.0,+1.9]$} \\
\bottomrule
\end{tabular}

\end{table}

\paragraph{Smarter voting widens the gap.} The ensemble's advantage over mixed-model debate is not an artifact of naive aggregation. Re-scoring every matched ensemble with votes weighted by each member's solo task accuracy widens its lead over debate everywhere tiers are mixed: from $-1.04$ to $-1.35$pp on core triples and from $-2.16$ to $-3.40$pp on pairs, where down-weighting the weak member seems to matter the most (Table~\ref{tab:weighted}). Seed-42 weights applied to seed-42 problems are in-sample; the key evidence is the seed-43 replication, where the same seed-42-derived weights are applied to unseen problems and reproduce the widened lead exactly ($-1.35$pp). Calibration must respect the task's choice count. \emph{Binary} log-odds weighting backfires ($-1.68$pp vs.\ plain majority, pooled) because members with solo accuracy below 50\% receive negative weights and anti-vote while still being far above 4--5-way chance. The correct multiclass form $\log\!\big(p(K{-}1)/(1{-}p)\big)$ repairs this: it beats debate significantly on core triples at both seeds ($-1.11$pp at seed 42) and turns the previously null quads significantly pro-ensemble ($-0.51$pp), while the pairs delta is directionally negative but not significant ($-1.59$ $[-3.12, +0.18]$; Table~\ref{tab:weighted}). Even the miscalibrated rule leaves debate no room and the main conclusion stands: deliberation never beats the best available aggregation of the same generations.

Table~\ref{tab:tiers} stratifies the core triples by capability-tier composition. Accuracy is monotone in composition for both arms because aggregate member capability, not heterogeneity, sets team performance. The significant deficits concentrate in mid-to-weak compositions; all-strong teams are the one composition where debate significantly edges its ensemble ($+0.44$).

Table~\ref{tab:rounds} varies the number of debate rounds over ten core teams. Essentially the entire gain arrives with the first exchange ($r{=}2$); rounds 3--5 are flat, and CSQA declines beyond $r{=}2$. The all-weak composition gains most from the first exchange ($61.5\to74.2$) and then erodes.

\begin{table}[H]\tiny\centering
\caption{Weighted-voting robustness of the mixed-model result. Each cell is
the paired delta debate $-$ ensemble (pp; negative favors the ensemble)
under three aggregation rules over the same stored generations: plain
majority, votes weighted linearly by each member's solo seed-42 accuracy on
the task, and Naive-Bayes log-odds weights $\log(p/(1-p))$ ($p$ clipped to
$[0.02, 0.98]$). 95\% CIs from a 10k bootstrap over teams; $^*$ = CI
excludes zero.}
\label{tab:weighted}
\begin{tabular}{lccccc}
\toprule
Team family & Teams & vs.\ majority & vs.\ acc.-weighted & vs.\ log-odds (binary) & vs.\ log-odds (multiclass) \\
\midrule
Core triples, seed 42 & 80 & $-1.04$ [$-1.49$, $-0.60$]$^*$ & $-1.35$ [$-1.81$, $-0.92$]$^*$ & $+0.63$ [$-0.25$, $+1.58$] & $-1.11$ [$-1.62$, $-0.59$]$^*$ \\
Core triples, seed 43 & 80 & $-0.93$ [$-1.38$, $-0.48$]$^*$ & $-1.35$ [$-1.80$, $-0.92$]$^*$ & $+0.57$ [$-0.27$, $+1.44$] & $-1.30$ [$-1.80$, $-0.83$]$^*$ \\
Pairs (AAB/ABB) & 30 & $-2.16$ [$-3.42$, $-0.88$]$^*$ & $-3.40$ [$-4.52$, $-2.33$]$^*$ & $+2.10$ [$-0.12$, $+4.48$] & $-1.59$ [$-3.12$, $+0.18$] \\
Quads (dup-s/w) & 40 & $+0.30$ [$-0.12$, $+0.74$] & $-0.35$ [$-0.74$, $+0.02$] & $+0.39$ [$-0.22$, $+1.09$] & $-0.51$ [$-0.89$, $-0.16$]$^*$ \\
\bottomrule
\end{tabular}

\end{table}

\begin{table}[H]\small\centering
\caption{Core-triple accuracy by capability-tier composition (S/M/W =
strong/mid/weak member tiers), debate vs.\ matched ensemble. $^*$: the
$\Delta$'s 95\% bootstrap CI over that composition's eight teams excludes
zero.}
\label{tab:tiers}
\begin{tabular}{lccc}
\toprule
Composition & Debate & Ensemble & $\Delta$ (pp) \\
\midrule
SSS & 93.7 & 93.2 & $+0.44$$^*$ \\
SSM & 92.5 & 92.8 & $-0.28$ \\
SMM & 90.3 & 91.5 & $-1.22$ \\
MMM & 87.0 & 88.2 & $-1.16$$^*$ \\
SSW & 91.7 & 92.2 & $-0.59$ \\
SMW & 90.6 & 90.9 & $-0.31$ \\
MMW & 84.9 & 87.1 & $-2.19$$^*$ \\
SWW & 87.4 & 89.3 & $-1.91$ \\
MWW & 80.8 & 83.7 & $-2.87$$^*$ \\
WWW & 75.6 & 76.7 & $-1.13$ \\
\bottomrule
\end{tabular}

\end{table}

\begin{table}[H]\small\centering
\caption{Rounds ablation: mixed-team debate accuracy (\%) by number of
rounds, pooled over ten core teams spanning every tier composition
(SSS--WWW). $r{=}1$ is a single-round majority vote over the team; each
added round costs $n$ further generations, so columns are not
budget-matched; the row reads as what additional rounds buy. $^*$: the
paired difference from $r{=}1$ over the ten teams has a 95\% bootstrap CI
excluding zero.}
\label{tab:rounds}
\begin{tabular}{lccccc}
\toprule
Task & $r{=}1$ & $r{=}2$ & $r{=}3$ & $r{=}4$ & $r{=}5$ \\
\midrule
Math & 93.5 & 98.3 & \textbf{99.2}$^*$ & \textbf{99.2} & \textbf{99.2} \\
GSM8K & 91.7 & 93.2 & 94.0 & 94.3 & \textbf{94.5} \\
MMLU & 74.1 & 76.4 & 76.8 & \textbf{77.0}$^*$ & 75.6 \\
CSQA & 79.2 & \textbf{81.0}$^*$ & 79.6 & 79.7 & 78.6 \\
\midrule
Mean & 84.6 & 87.2$^*$ & 87.4$^*$ & 87.5$^*$ & 87.0$^*$ \\
\bottomrule
\end{tabular}

\end{table}

\section{Biography Evaluation}
\label{app:biography}
\label{app:judges}

Because no automatic scorer exists for the biography task, judging is done by a panel of LLM judges chosen to be vendor-disjoint from all participants: Gemma-3-27B (Google), GPT-OSS-20B, and GPT-OSS-120B (OpenAI), with a fourth judge serving only a coupling ablation (Table~\ref{tab:judges}). Because judges often disagree, it is important to look at aggregation results. The two OpenAI judges are near-duplicates (per-cell persona effects correlate at $r{=}0.95$), so a plain three-judge mean would weight OpenAI two-to-one over Google; the vendor-balanced score therefore averages the OSS pair into one vote before averaging with Gemma.

The sampling control's margin comes partly from pooling mechanics. The control pools the nine biographies' claims before scoring: precision is computed over the \emph{union} of extracted claims per subject, not averaged per run, so more verifiable claims accumulate per subject and single unrepresentative drafts wash out. The control's $+10.0$pp margin over solo splits into $+2.0$pp from temperature spread and $+8.0$pp from multiplicity, the latter partly reflecting eight models whose single solo biography is unrepresentative; the debate comparison, which pools three biographies per subject, is the robust one.

Claim precision does not penalize epistemic refusal: a model that declines subjects it does not know (the LFM family, on up to 10 of 40 subjects in some cells) is judged only on what it writes, flattering honest abstainers over confabulators. This is partly the reason why the smallest LFM model tops the per-model table below. The persona penalty is judge-vendor-dependent, which we report for full transparency: Gemma alone sees a uniformly significant penalty ($-7.4$pp, $p<10^{-4}$), while either OpenAI judge alone sees none. Cross-vendor agreement is moderate while intra-vendor agreement is near-perfect: the GPT-OSS-20B/120B pair agrees at $\kappa$ 0.759/0.762 (claim and fact directions), exceeding every cross-vendor pair (0.32--0.57), so a $6\times$ difference in judge scale alters agreement far less than a change of vendor (Table~\ref{tab:kappa}). The disagreement has a direction: the Google judge is systematically more generous than either OpenAI judge. The supplementary judge, Muse Glimmer, fully shares a vendor with two participant models; it serves only the judge-coupling ablation and is never pooled into headline scores (Table~\ref{tab:judges}).

\begin{table}[H]\small\centering
\caption{Per-model biography factuality: claim precision among verifiable
claims (\%) under the vendor-balanced three-judge panel (one vote per judge
vendor; Gemma-3-27B vs.\ the averaged GPT-OSS-20B/120B pair), seed 42,
40 subjects, ctx-max policy. Solo: single agent. Debate: 3 agents $\times$
3 rounds. SC-9: nine independent single-round biographies pooled through
the same panel. Persona: maximally-diverse (MaxDet) persona team, 3
agents. ``Verif.\ claims'': mean verifiable claims per cell under the
debate arm --- a volume indicator (precision is computed only over these).
Bold: best per row.}
\label{tab:biography_permodel}
\begin{tabular}{lccccc}
\toprule
Model & Solo & Debate & SC-9 & Persona & Verif.\ claims \\
\midrule
LFM2.5-1.2B & 87.3 & 86.5 & 86.3 & \textbf{92.6} & 224 \\
LFM2.5-8B-A1B & 40.8 & 42.9 & \textbf{44.1} & 32.5 & 340 \\
Llama-3.2-3B & \textbf{30.0} & 28.4 & 28.0 & 27.1 & 389 \\
Llama-4-Scout-17B-16E-Ins & \textbf{69.9} & 62.0 & 69.0 & 58.7 & 490 \\
Ministral-3-8B-2 & 31.4 & 55.9 & \textbf{64.7} & 16.7 & 865 \\
Mistral-7B-v0.3 & 38.7 & 40.9 & \textbf{52.1} & 26.1 & 464 \\
NVIDIA-Nemotron-3-Nano-30 & 45.4 & 57.7 & 60.6 & \textbf{63.6} & 441 \\
NVIDIA-Nemotron-3-Nano-4B & 19.1 & 19.2 & \textbf{34.2} & 21.2 & 470 \\
Olmo-3-7B & 49.4 & 53.9 & \textbf{65.7} & 61.0 & 238 \\
Phi-4-mini & 36.4 & 44.2 & \textbf{61.5} & 34.9 & 426 \\
Qwen3-0.6B & 47.1 & 31.9 & \textbf{51.6} & 19.2 & 147 \\
Qwen3-14B & 73.9 & 66.0 & \textbf{81.9} & 65.0 & 387 \\
Qwen3-32B (W4A16) & 54.9 & 65.6 & \textbf{81.2} & 55.5 & 340 \\
Qwen3.5-27B (GPTQ) & 64.5 & 67.2 & \textbf{72.4} & 56.8 & 2389 \\
Qwen3.5-35B-A3B (GPTQ) & 64.0 & 65.3 & \textbf{73.3} & 54.4 & 2966 \\
Qwen3.5-9B & 54.9 & 55.8 & \textbf{62.7} & 45.8 & 2791 \\
R1-Distill-Qwen-14B & 74.9 & 71.0 & \textbf{77.8} & 74.3 & 721 \\
R1-Distill-Qwen-32B-w4a16 & 75.3 & 77.7 & \textbf{86.5} & 77.5 & 911 \\
VibeThinker-1.5B & 18.3 & 10.4 & \textbf{20.8} & 10.9 & 368 \\
gemma-4-12B-it & 53.1 & 58.7 & 73.9 & \textbf{75.3} & 465 \\
gemma-4-26B-A4B-it-FP8-dy & 77.6 & 76.2 & \textbf{87.2} & 74.7 & 457 \\
granite-4.1-8b & 75.7 & 73.2 & 76.4 & \textbf{77.5} & 333 \\
phi-4 & 72.8 & 68.9 & \textbf{73.6} & 71.2 & 944 \\
\midrule
Mean & 54.6 & 55.6 & 64.6 & 51.8 & 764 \\
\bottomrule
\end{tabular}

\end{table}

\begin{table}[H]\small\centering
\caption{The four-judge biography panel over every judged file (mirror-
deduped). ``Strict'': supported / all claims; ``Verif.'': supported /
(supported + contradicted). The three primary judges sit above the rule;
the supplementary coupling-ablation judge (Muse Glimmer) below.}
\label{tab:judges}
\begin{tabular}{lrccrcc}
\toprule
& \multicolumn{3}{c}{Claims direction} & \multicolumn{3}{c}{Facts direction} \\
\cmidrule(lr){2-4}\cmidrule(lr){5-7}
Judge & Units & Strict & Verif. & Units & Strict & Verif. \\
\midrule
Gemma-3-27B & 6.98M & 18.4 & 65.2 & 1.91M & 39.8 & 78.9 \\
GPT-OSS-20B & 6.95M & 11.2 & 58.6 & 1.90M & 9.7 & 57.9 \\
GPT-OSS-120B & 6.98M & 11.1 & 61.6 & 1.91M & 9.9 & 52.2 \\
\midrule
Muse Glimmer & 3.30M & 10.3 & 42.2 & 0.96M & 12.7 & 42.1 \\
\bottomrule
\end{tabular}

\end{table}

\begin{table}[H]\small\centering
\caption{Pairwise Cohen's $\kappa$ between judges over record-aligned
verdicts ($3.3$--$7.0$M claim / $1.0$--$1.9$M fact pairs per judge pair): claims
direction below the diagonal, facts direction above.}
\label{tab:kappa}
\begin{tabular}{lcccc}
\toprule
& Gemma-3-27B & GPT-OSS-20B & GPT-OSS-120B & Muse Glimmer \\
\midrule
Gemma-3-27B & --- & 0.342 & 0.372 & 0.317 \\
GPT-OSS-20B & 0.563 & --- & 0.762 & 0.440 \\
GPT-OSS-120B & 0.568 & 0.759 & --- & 0.474 \\
Muse Glimmer & 0.354 & 0.414 & 0.412 & --- \\
\bottomrule
\end{tabular}

\end{table}

\section{The Context-Window Curve}
\label{app:window}

\begin{figure}[H]
\centering
\includegraphics[width=0.62\linewidth]{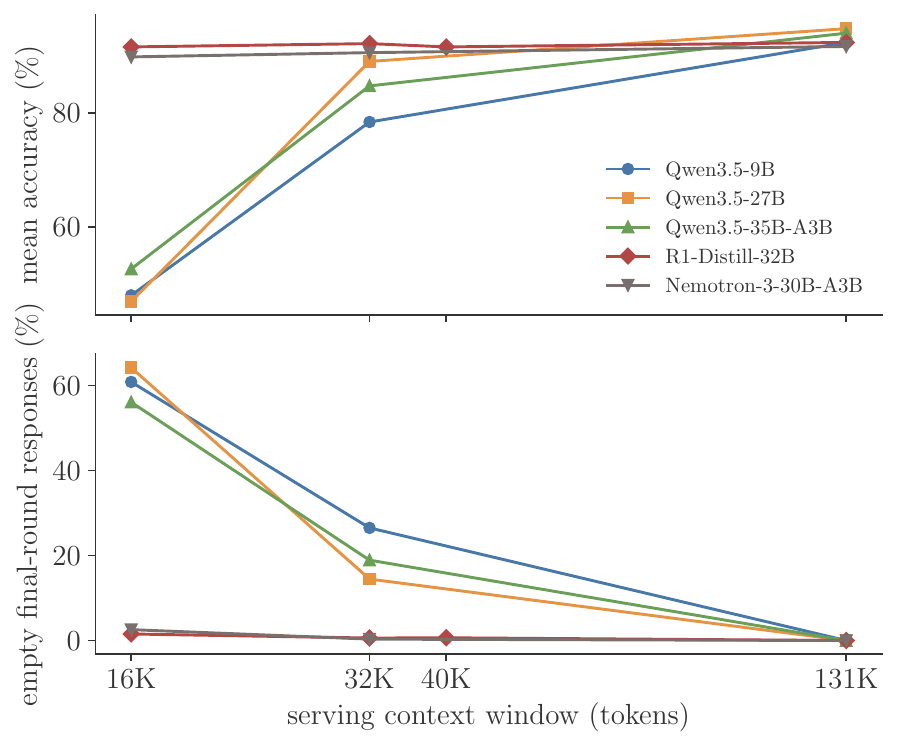}
\caption{Debate vs.\ the serving context window for the five
budget-exceeding models (cells paired within model). Top: mean accuracy,
inline-scored tasks; bottom: fraction of final-round responses left empty
by transcript overflow.}
\label{fig:window_curve}
\end{figure}

Figure~\ref{fig:window_curve} traces debate accuracy against the served context window. The Qwen3.5 thinkers lose half their accuracy at 16K and recover monotonically to 131K as empty final rounds vanish; the two models whose budgets fit their windows are flat across the sweep. The effect is a serving artifact rather than a change in ability, which is what motivates the context-corrected numbers used throughout the paper.

\section{Benchmark Slices: MMLU-Pro, GSM-Symbolic, and GPQA Diamond}
\label{app:p05}
\label{sec:headroom}

A fair objection to any debate study is that its benchmarks are too easy or too memorized for the mechanism to show. Three targeted slices address this concern. We confirmed that headroom does exist below our grid: on MMLU-Pro the same 23 models score $\sim$15 points below plain MMLU (all 23 drop). Contamination does not appear to drive the results: on GSM-Symbolic, whose templates perturb GSM8K surface forms, every configuration's results are within a point of its GSM8K counterpart, and debate and sampling are at parity ($+0.35$pp; per-model data in Table~\ref{tab:p05}).

The hardest slice closes what looked like the study's one exception. In a seven-model pilot experiment, debate led its matched sampling control on GPQA Diamond by four points. Scaling to all 23 models shrinks the lead to $+3.2$pp, not statistically significant, and correcting a second silent budget artifact eliminates it. Our engine assigned the full 40{,}960-token thinking budget by matching model names, a pattern that failed to recognize the Qwen3 and Nemotron reasoning families and left them at the 4{,}096-token default. On questions this hard, capped single-shot reasoning runs off the budget mid-derivation, while debate's later rounds anchor on peers' answers, arriving at a conclusion early: the cap handicapped the \emph{control}. After re-running the experiments at the full budget, the pooled comparison lands at solo $49.1$, debate $54.3$, SC-9 $54.9$: debate $-$ SC-9 $=-0.6$pp $[-2.2, +1.0]$, null like everywhere else. The budget cap is context overflow's twin (Section~\ref{sec:ctx}): both silently handicap one arm of a matched comparison, and both must be audited before deliberation is credited.

What survives after all these experiments is the ensemble effect. Debate still beats \emph{solo} here by $+5.2$pp, but so does sampling, and the weak-duplicated-quad MMLU cell (Table~\ref{tab:mixed_bytask}) remains the sole positive for deliberation; hardness alone is clearly not where debate pays (sampling edges debate on MMLU-Pro too, by $2.0$pp mean over all 23 models). This is also why the small-model setting is the one in which the question is answerable at all: a frontier model near ceiling leaves debate nothing to improve, and the comparison must be affordable enough to run thousands of matched controls.

\begin{table}[H]\small\centering
\caption{Per-model accuracy (\%) on the headroom slice (MMLU-Pro) and the
contamination slice (GSM-Symbolic): Solo, Debate ($3\times3$), and SC-9,
seed 42, 100 problems, all 23 models.}
\label{tab:p05}
\begin{tabular}{lcccccc}
\toprule
 & \multicolumn{3}{c}{MMLU-Pro} & \multicolumn{3}{c}{GSM-Symbolic} \\
\cmidrule(lr){2-4}\cmidrule(lr){5-7}
Model & Solo & Debate & SC-9 & Solo & Debate & SC-9 \\
\midrule
gemma-4-12B-it & 74.0 & 77.0 & 78.0 & 96.0 & 95.0 & 96.0 \\
gemma-4-26B-A4B-it-FP8-dy & 82.0 & 85.0 & 83.0 & 98.0 & 97.0 & 98.0 \\
granite-4.1-8b & 43.0 & 56.0 & 59.0 & 95.0 & 96.0 & 97.0 \\
LFM2.5-1.2B & 32.0 & 35.0 & 35.0 & 77.0 & 83.0 & 74.0 \\
LFM2.5-8B-A1B & 40.0 & 60.0 & 65.0 & 93.0 & 94.0 & 94.0 \\
Llama-3.2-3B & 11.0 & 17.0 & 37.0 & 52.0 & 71.0 & 83.0 \\
Llama-4-Scout-17B-16E-Ins & 70.0 & 70.0 & 71.0 & 97.0 & 99.0 & 99.0 \\
Ministral-3-8B-Instruct-2 & 58.0 & 70.0 & 69.0 & 95.0 & 95.0 & 97.0 \\
Mistral-7B-Instruct-v0.3 & 19.0 & 20.0 & 31.0 & 27.0 & 52.0 & 29.0 \\
NVIDIA-Nemotron-3-Nano-30 & 73.0 & 77.0 & 83.0 & 94.0 & 95.0 & 96.0 \\
NVIDIA-Nemotron-3-Nano-4B & 58.0 & 62.0 & 66.0 & 96.0 & 98.0 & 96.0 \\
Olmo-3-7B & 50.0 & 56.0 & 55.0 & 90.0 & 96.0 & 94.0 \\
phi-4 & 75.0 & 76.0 & 72.0 & 98.0 & 99.0 & 99.0 \\
Phi-4-mini & 38.0 & 43.0 & 48.0 & 80.0 & 87.0 & 92.0 \\
Qwen3-0.6B & 27.0 & 34.0 & 39.0 & 82.0 & 85.0 & 86.0 \\
Qwen3-14B & 72.0 & 75.0 & 79.0 & 97.0 & 97.0 & 96.0 \\
Qwen3-32B-quantized.w4a16 & 72.0 & 81.0 & 80.0 & 97.0 & 98.0 & 98.0 \\
Qwen3.5-27B-GPTQ-Int4 & 78.0 & 53.0 & 70.0 & 96.0 & 83.0 & 92.0 \\
Qwen3.5-35B-A3B-GPTQ-Int4 & 79.0 & 82.0 & 71.0 & 96.0 & 98.0 & 96.0 \\
Qwen3.5-9B & 72.0 & 76.0 & 61.0 & 98.0 & 97.0 & 90.0 \\
R1-Distill-Qwen-14B & 59.0 & 71.0 & 71.0 & 96.0 & 97.0 & 99.0 \\
R1-Distill-Qwen-32B-w4a16 & 62.0 & 73.0 & 71.0 & 99.0 & 98.0 & 99.0 \\
VibeThinker-1.5B & 32.0 & 49.0 & 50.0 & 94.0 & 94.0 & 96.0 \\
\midrule
Mean & 55.5 & 60.8 & 62.8 & 88.8 & 91.5 & 91.1 \\
\bottomrule
\end{tabular}

\end{table}

\paragraph{GPQA Diamond forensics.} Eight reasoning models missed by the engine's name-matching pattern ran at a 4{,}096-token thinking budget, and the cap punished the single-shot control hardest: SC-9 parse rates fell as low as $60\%$ as capped reasoning ran off the budget mid-derivation. Re-run at the full 40{,}960-token budget, the eight capped models' SC-9 cells gain up to $+34$pp; 19 of 23 per-model debate$-$SC-9 deltas are individually null and the four significant ones split two per side. Realized token accounting confirms the correction unbinds the control rather than favoring either arm: liberation raises SC-9's generated tokens more than debate's (median $1.8\times$ vs.\ $1.4\times$), and under corrected budgets debate generates a median $0.6\times$ SC-9's completion tokens while tying the comparison, because later rounds anchor on peers' answers and conclude early. The corrected runs also expose a protocol ceiling: with three 40K-budget thinking agents, round-3 prompts reach 100--112K tokens and generation is clamped to fit the 131{,}072-token serving window. At three rounds, a 27B-class thinker fills the largest window we can serve.

\section{Wall-Clock and Token Cost}
\label{app:cost}

Table~\ref{tab:cost} gives the realized cost accounting. Generated-token ratios near $1\times$ confirm the budgets are matched. Debate re-reads its peers' growing transcript at every round, so its \emph{prompt}-side consumption is roughly ten times the control's and its total token bill runs $2.3$--$4.1\times$ across tasks, while losing or tying on accuracy (Table~\ref{tab:main}). Persona debate adds a further $1.26\times$ wall-clock over plain debate.

\begin{table}[H]\small\centering
\caption{Mean wall-clock minutes per 100-problem job (biography: 40
subjects) on the 2$\times$RTX 3090 workstation, pooled over the 23-model
pool at seed 42, plus token-level ratios computed by re-tokenizing every
stored conversation with each model's own tokenizer (median over the 23
models' complete debate/SC file pairs).}
\label{tab:cost}
\begin{tabular}{lccccccc}
\toprule
 & \multicolumn{4}{c}{wall-clock (min)} & \multicolumn{3}{c}{debate/SC-9 ratio} \\
\cmidrule(lr){2-5}\cmidrule(lr){6-8}
Task & Solo & Debate & SC-9 & Persona & clock & gen.\ tokens & all tokens \\
\midrule
Math & 4 & 11 & 8 & 14 & 1.34$\times$ & 1.18$\times$ & 2.34$\times$ \\
GSM8K & 6 & 23 & 18 & 28 & 1.32$\times$ & 0.84$\times$ & 3.36$\times$ \\
MMLU & 7 & 27 & 16 & 34 & 1.74$\times$ & 0.93$\times$ & 3.51$\times$ \\
CSQA & 5 & 21 & 12 & 30 & 1.82$\times$ & 1.10$\times$ & 3.80$\times$ \\
Biography & 4 & 17 & 9 & 19 & 1.95$\times$ & 1.26$\times$ & 4.09$\times$ \\
\midrule
Mean & 5 & 20 & 12 & 25 & 1.61$\times$ & 1.06$\times$ & 3.42$\times$ \\
\bottomrule
\end{tabular}

\end{table}

\section{Protocol Details}
\label{app:protocol}

This appendix records the protocol constants a reimplementation needs. All of them are fixed across every arm.

The debate prompts follow \citet{du2023improving}. Round 1 poses the task question directly. Later rounds prepend ``These are the solutions to the problem from other agents:'', followed by each peer's full previous-round response delimited in triple backticks, then ``Using the solutions from other agents as additional information, can you provide your answer\ldots'', with the task-specific answer-format instruction after it. The single-agent $n{=}1$ arm instead receives ``Can you double check that your answer is correct\ldots''.

Majority-vote ties are broken by the first-encountered answer among parseable responses, identically for debate and sampling arms. Benchmark subsampling is uniform at random without subject stratification, drawn once per task with fixed seed 42 (seed 43 for the replication slice) and reused across all arms, so every comparison is paired on identical problems. Capability tiers partition the pool by mean single-agent accuracy over GSM8K, MMLU, and CSQA: strong $\ge 0.85$, mid $0.60$--$0.85$, weak $<0.60$.

Biography claims are extracted by a deterministic parser that segments the generated biography at bullet points; no LLM takes part in the extraction step, and judges classify each extracted claim against the reference article. Determinant computation uses the exact sign-log-determinant. The smallest selected-team determinant observed across all models is $2.7\times10^{-10}$, far above underflow.

\section{Diversity-Metric Robustness}
\label{app:metricrobust}

The persona-diversity ladder is not a log-det artifact. Table~\ref{tab:metricrobust} re-ranks every candidate trio under the max--min pairwise-distance objective and compares the two orderings per model. Every re-extraction reproduces the published MaxDet team as the combo-space argmax, a check asserted programmatically in the released pipeline. Under both metrics, in all 23 models, the most-redundant teams remain the most redundant and the MaxDet teams remain near-optimal.

\begin{table}[H]\small\centering
\caption{Robustness metrics of persona diversity. Per model:
Spearman correlation of MaxMin (minimum pairwise cosine distance) vs.\
log-det over all $\binom{50}{3}=19{,}600$ candidate trios (``combo
$\rho$''); Spearman of the ladder teams' MaxMin percentiles against rung
(``ladder $\rho$''); and the percentile of the rung-0 and MaxDet teams
under the MaxMin ranking.}
\label{tab:metricrobust}
\begin{tabular}{lcccc}
\toprule
Model & combo $\rho$ & ladder $\rho$ & rung-0 pctl & MaxDet pctl \\
\midrule
gemma-4-12B-it & 0.814 & 0.794 & 0.18 & 99.86 \\
gemma-4-26B-A4B-it-FP8-dyn & 0.835 & 0.964 & 0.12 & 99.96 \\
granite-4.1-8b & 0.895 & 0.964 & 0.39 & 99.98 \\
LFM2.5-1.2B-Instruct & 0.923 & 0.988 & 0.05 & 99.82 \\
LFM2.5-8B-A1B & 0.877 & 0.818 & 0.05 & 100.00 \\
Llama-3.2-3B-Instruct & 0.893 & 0.879 & 0.45 & 99.99 \\
Llama-4-Scout-17B-16E-w4a16 & 0.867 & 0.782 & 0.10 & 100.00 \\
Ministral-3-8B-Instruct-2512 & 0.906 & 0.867 & 0.21 & 99.73 \\
Mistral-7B-Instruct-v0.3 & 0.895 & 0.915 & 0.12 & 99.99 \\
Nemotron-3-Nano-30B-A3B-FP8 & 0.826 & 0.915 & 0.35 & 100.00 \\
Nemotron-3-Nano-4B & 0.892 & 0.964 & 0.23 & 100.00 \\
Olmo-3-7B-Instruct & 0.874 & 0.818 & 1.07 & 99.99 \\
phi-4 & 0.829 & 0.952 & 0.37 & 100.00 \\
Phi-4-mini-instruct & 0.847 & 0.673 & 0.48 & 99.75 \\
Qwen3-0.6B & 0.869 & 0.830 & 0.12 & 100.00 \\
Qwen3-14B & 0.889 & 0.927 & 0.03 & 99.79 \\
Qwen3-32B-quantized.w4a16 & 0.837 & 0.830 & 0.20 & 99.97 \\
Qwen3.5-27B-GPTQ-Int4 & 0.894 & 0.952 & 0.03 & 99.99 \\
Qwen3.5-35B-A3B-GPTQ-Int4 & 0.898 & 0.964 & 0.08 & 99.99 \\
Qwen3.5-9B & 0.891 & 0.867 & 0.01 & 100.00 \\
R1-Distill-Qwen-14B & 0.870 & 0.818 & 0.44 & 100.00 \\
R1-Distill-Qwen-32B-w4a16 & 0.845 & 0.915 & 0.03 & 100.00 \\
VibeThinker-1.5B & 0.892 & 0.903 & 0.35 & 100.00 \\
\midrule
Median & 0.877 & 0.903 & 0.18 & 99.99 \\
\bottomrule
\end{tabular}

\end{table}

\section{Practical Implications}
\label{app:practical}

For practitioners: at fixed budget, prefer sampling-plus-voting over debate (equal or better accuracy at $1.6\times$ less wall-clock); pool \emph{different} small models and vote (mixed rosters lift weak deployments 10--20 points over their best member, no debate needed); if debating, two rounds suffice; do not assign personas to small models; size the context window to agents $\times$ rounds $\times$ per-turn budget; and judge long-form output with a vendor-balanced panel (judge vendor dominated evaluation variance, dwarfing a $6\times$ judge-scale difference).

For researchers proposing richer debate mechanisms (process-level critique, verification tools, dynamic path allocation~\citep{li2026dynadebate}), our grid doubles as a stringent, reproducible baseline: under matched generation budgets, and more strictly still under token-level accounting (debate additionally pays to re-read peer transcripts each round; Appendix~\ref{app:cost}), standard answer-exchange debate does not clear sampling-plus-voting.

\section{The Persona Set}
\label{app:personas}

The 50 personas below are the complete candidate pool from which every
persona team in the paper is selected (MaxDet-optimal teams, decile-ladder
rungs, and the minimum-determinant teams alike). Each is injected verbatim
as the agent's system prompt in the form ``You are \{description\}.''
{\small
\begin{longtable}{@{}p{0.16\linewidth}p{0.78\linewidth}@{}}
\toprule
Callsign & Description \\ \midrule
\endfirsthead
\toprule
Callsign & Description \\ \midrule
\endhead
\bottomrule
\endlastfoot
cryptographer & a nihilistic cryptographer who only trusts solutions verifiable by zero-knowledge proofs \\
baroque & a Baroque music theorist fixated on harmonic counterpoint and structural symmetry \\
meteorologist & a chaotic systems meteorologist who views all certainty as a transient statistical anomaly \\
xenolinguist & a hard-science fiction xenolinguist obsessed with logical consistency and species-specific syntax \\
cartographer & a medieval cartographer whose primary concern is establishing clear boundaries and known territories \\
neuropharm & a neuropharmacologist who analyzes all input as complex patterns of reward and aversion chemicals \\
minimalist & an extreme minimalist architect whose goal is to strip the solution down to its bare, essential structure \\
quantum & a quantum physicist who models all decisions as probabilistic wave function collapse \\
jurist & a Roman imperial jurist who strictly adheres to precedent and codified legal language \\
urbanplanner & an early 20th-century urban planner focused on maximizing geometric efficiency and transit flow \\
anarchist & a radical anarchist who views all imposed structures and hierarchies as fundamentally flawed \\
utilitarian & a utilitarian extremist who focuses solely on maximizing the benefit for the greatest number, regardless of individual cost \\
pessimist & a Schopenhauerian pessimist who assumes the most detrimental outcome is inevitable and prepares for it \\
deontologist & a Kantian deontologist who judges all actions strictly by their moral imperative and universal rule application \\
contraskeptic & a contrarian skeptic whose sole purpose is to argue the inverse of the majority opinion at all times \\
romantic & an unfettered Romantic idealist who prioritizes artistic vision and emotional resonance over cold logic \\
capitalist & a hyper-capitalist financier who evaluates every move solely on ROI (Return on Investment) and marginal profit \\
materialist & a pre-Socratic materialist who reduces all problems to their basic physical components and forces \\
taoist & a Taoist sage who seeks the path of least effort and accepts paradoxical outcomes \\
solipsist & a solipsistic egoist who only values input that directly confirms or serves their own internal world view \\
enigma & an enigma machine operator whose primary filter is signal-to-noise ratio and encrypted hidden messages \\
gothic & a Gothic novelist who focuses on dramatic irony, foreshadowing, and latent horror \\
prankster & a childish prankster whose motivation is to introduce creative chaos and subvert expectations \\
victorian & a Victorian etiquette consultant obsessed with proper formatting, deference, and rigid social boundaries \\
bureaucrat & a Soviet-era bureaucrat who prioritizes documentation, adherence to arbitrary quotas, and triplicate forms \\
infant & a pre-lingual infant whose understanding is limited to basic sensations, needs, and spatial presence \\
deconstructionist & a post-modern deconstructionist who questions the validity and inherent meaning of all proposed terms \\
pirate & a dread pirate captain whose strategy is based on risk assessment, immediate plunder, and intimidation \\
zenmaster & a Zen master who communicates only through non-sequiturs, koans, and minimal, cryptic statements \\
adman & a 1950s Madison Avenue ad man who views the solution as a campaign to be emotionally sold and packaged \\
volcanologist & a deep-sea volcanologist focused on extremes of pressure, heat, and slow geologic change \\
pathologist & a forensic pathologist who works backward from the failure state to determine the precise cause of death \\
cosmichorror & a cosmic horror narrator who frames the problem as an ancient, unknowable, and terrifying truth \\
syseng & a systems engineer who focuses strictly on modularity, inter-component dependencies, and error states \\
comedian & a stand-up comedian who evaluates suggestions based on their absurdity, timing, and satirical value \\
auditor & a resource auditor who views every input as a budget line item that must be strictly justified \\
sidekick & a mythological hero's sidekick who is overly cautious, risk-averse, and constantly points out dangers \\
grandmaster & an expert chess grandmaster who analyzes all moves based on look-ahead, board state, and counterplay \\
painter & a Renaissance painter who values perspective, light, shadow, and visual harmony in the final presentation \\
inspector & an industrial safety inspector whose primary filter is identifying catastrophic single points of failure \\
revisionist & a historical revisionist who assumes all primary sources are propaganda and seeks hidden motives \\
dronecommander & a drone swarm commander who views the task as parallel processing over numerous, interchangeable units \\
psychologist & a child psychologist who interprets all actions through the lens of developmental stage and emotional need \\
alchemist & a hermetic alchemist who seeks to transmute the problem into a perfect, pure, and philosophical gold standard \\
wrestler & a professional wrestler whose analysis focuses on dramatic confrontation, stage presence, and audience reaction \\
gambler & a street-smart gambler whose strategy is based on calculating odds, weighted risk, and bluffing potential \\
survivalist & a hermitic survivalist who prioritizes self-sufficiency, redundancy, and defense against external threats \\
gossip & a city gossip columnist who filters information based on intrigue, conflict, and personal scandal potential \\
scavenger & a post-apocalyptic scavenger who judges solutions only by their immediate utility, robustness, and salvageability \\
anthropologist & a linguistic anthropologist who views the solution as a structure of cultural symbols and shared meaning \\
\end{longtable}

}

\end{document}